\documentclass[letterpaper,journal]{IEEEtran}
\usepackage{amsmath,amsfonts}
\usepackage{algorithmic}
\usepackage{algorithm}
\usepackage{array}
\usepackage{textcomp}
\usepackage{stfloats}
\usepackage{url}
\usepackage{verbatim}
\usepackage{graphicx}
\usepackage{cite}
\usepackage{color}
\usepackage{mysymbol}
\usepackage{booktabs}
\usepackage{multirow}
\usepackage{enumitem}
\usepackage{subfigure}
\usepackage{hyperref}
 
\begin{document}

\title{Explainable Spatio-Temporal GCNNs  for Irregular Multivariate Time Series: Architecture and Application to ICU Patient Data}

\author{Óscar Escudero-Arnanz,~\IEEEmembership{Student Member,~IEEE},
Cristina Soguero-Ruiz,
Antonio G. Marques,~\IEEEmembership{Senior Member,~IEEE}
\vspace{-0.5cm}
\thanks{All the authors are with the department of Signal Processing at King Juan Carlos University, Spain. This work was supported by the Spanish AEI (DOI 10.13039/501100011033) under Grants PID2022-136887NBI00 and PID2023-149457OB-I00, and by the Autonomous Community of Madrid within the ELLIS Unit Madrid framework.}}
 


\maketitle

\begin{abstract}
In this paper, we present XST-GCNN (eXplainable Spatio-Temporal Graph Convolutional Neural Network), an innovative architecture designed for processing heterogeneous and irregular Multivariate Time Series (MTS) data. Our processing architecture captures both temporal and feature dependencies within a unified spatio-temporal pipeline by leveraging a GCNN that uses a spatio-temporal graph and aims at optimizing predictive accuracy and interpretability.
For graph estimation, we propose several techniques, including a novel approach based on the (heterogeneous) Gower distance. Once the graphs are estimated, we propose two approaches for graph construction: one based on the Cartesian product that treats temporal instants homogeneously, and a spatio-temporal approach that considers different graphs per time step. 
Finally, we propose two GCNN architectures: a standard GCNN with a normalized adjacency matrix, and a higher-order polynomial GCNN. In addition to accuracy, we emphasize explainability by designing an inherently interpretable architecture and conducting a thorough interpretability analysis, identifying key feature-time combinations that drive the model's predictions.
We evaluate XST-GCNN using real-world Electronic Health Record data from the University Hospital of Fuenlabrada to predict Multidrug Resistance (MDR) in Intensive Care Unit patients, a critical healthcare challenge associated with high mortality and complex treatments. Our architecture outperforms traditional models, achieving a mean Receiver Operating Characteristic Area Under the Curve score of $\textbf{81.03} \pm \textbf{2.43}$. Additionally, the interpretability analysis provides actionable insights into clinical factors driving MDR predictions, enhancing model transparency and trust. This work sets a new benchmark for addressing complex inference tasks with heterogeneous and irregular MTS, offering a versatile and interpretable solution for real-world applications.
\end{abstract}


\section{Introduction}
\label{sec:Introduction}

Graphs have become powerful tools in both Machine Learning (ML) and Signal Processing (SP) due to their capacity to model complex interactions and capture intrinsic relationships within structured data~\cite{zhang2024graph}. In real-world applications, data often originate from multiple domains and exhibit heterogeneous characteristics, presenting a significant challenge for graph analysis and estimation~\cite{zhou2020graph, zhang2024graph}. This heterogeneity spans various data types—including numerical, categorical, and textual, among others—collected at different temporal and spatial scales, further complicating traditional graph construction methods~\cite{zhou2020graph}. These methods frequently rely on domain-specific adjustments for each variable or data source, leading to inconsistent graph representations and limiting their generalizability~\cite{zhou2020graph, wu2020comprehensive}.

To address the complexities described above, graph learning techniques have emerged as a powerful approach, allowing the inference of graph topologies directly from data, without imposing prior assumptions on the graph structure~\cite{zhang2021graph, gilmer2017neural}. However, conventional methods that construct multiple graphs for different data characteristics introduce redundancy and computational inefficiencies~\cite{isufi2021edgenets}. This highlights the need for advanced models capable of estimating a unified graph that integrates relationships among heterogeneous variables while maintaining computational efficiency and interpretability~\cite{velickovic2017graph}.

Once a unified graph that captures the relationships within heterogeneous data is estimated, it becomes foundational for subsequent inference and prediction tasks. Graph Convolutional Neural Networks (GCNNs) have proven highly effective in this context, leveraging the graph structure to capture hidden patterns by iterating over nodes and aggregating information from their neighbors~\cite{gama2020graphs}. More recent developments, such as Spatio-Temporal Graph Neural Networks (ST-GNNs), combine the strengths of GCNNs with Recurrent Neural Networks (RNNs) to handle both spatial and temporal dependencies in dynamic data~\cite{yu2018spatio, wu2020comprehensive, ruiz2020gated}. These models have demonstrated significant potential in solving complex problems involving heterogeneous and dynamic datasets~\cite{yu2018spatio}.

In response to the challenges of heterogeneous temporal data, we propose the eXplainable Spatio-Temporal Graph Convolutional Neural Network (XST-GCNN). This architecture is designed to efficiently capture and integrate Spatio-Temporal (ST) relationships within irregular Multivariate Time Series (MTS). Unlike conventional models, XST-GCNN unifies discrete and continuous data types into a cohesive graph representation, allowing for the modeling of complex, domain-spanning interactions. This approach not only estimates a single graph that encapsulates both spatial and temporal dependencies but also provides an interpretable architecture, essential for clinical decision-making. By capturing both local and global relationships within medical data, XST-GCNN offers a robust solution for enhancing outcome prediction and supporting decisions in high-stakes environments such as healthcare.

The effectiveness of the XST-GCNN architecture is demonstrated through a case study on real-world medical data from the Intensive Care Unit (ICU) of the University Hospital of Fuenlabrada (UHF), addressing the critical issue of Multidrug Resistance (MDR). Recognized by the World Health Organization as a growing global threat, MDR complicates infection treatments, increases mortality rates, and imposes significant pressures on healthcare systems~\cite{AMR_deaths_2023, whoAntimicrobialResistance_2023, world2024bacterial}. It is important to highlight that MDR is a particularly alarming subset of Antimicrobial Resistance (AMR).
The dataset, derived from Electronic Health Records (EHRs) of the ICU-UHF, was modeled as irregular MTS and included heterogeneous features such as real-valued and discrete variables. EHRs are frequently used to address a range of healthcare challenges, including early detection of sepsis, prediction of patient deterioration, and personalized treatment planning~\cite{shickel2017deep}. The inherent irregularity and variability in EHR data, recorded at varying intervals across different patients, exacerbate the limitations of traditional Time Series (TS) models, which struggle to provide accurate and interpretable predictions in clinical settings~\cite{wu2020comprehensive, xie2022deep}. Graph-based models, on the other hand, excel in representing the intricate relationships within clinical data as networks, facilitating more comprehensive analysis and better clinical insights~\cite{zhang2021graph}.

By capturing detailed, interconnected relationships within biomedical data, graph-based models enhance the integration and analysis of critical healthcare information, such as patient histories, diagnoses, and treatments~\cite{kallipolitis2023medical, zhang2021graph}. When combined with advanced SP techniques, these models significantly improve the extraction of meaningful patterns from clinical data, boosting both interpretability and predictive accuracy~\cite{gilmer2017neural, murali2023towards}. GCNNs, in particular, have revolutionized medical informatics by leveraging graph structures to reveal hidden patterns and improve the interpretability of clinical data~\cite{
scarselli2008graph}. This is especially valuable for assessing patient conditions and predicting clinical outcomes. Despite their potential, current research in GCNNs often overlooks the unique challenges posed by irregular MTS, such as varying recording frequencies and the heterogeneous nature of clinical data.

The XST-GCNN architecture proposed in this paper directly addresses these challenges (heterogeneity, irregular MTS, spatial and temporal dependencies, and the need for interpretability) by integrating spatial and temporal dependencies within heterogeneous clinical data, offering a unified, interpretable architecture that enhances clinical decision-making in the context of MDR.

\subsection{Related Work}
\label{sec:RelatedWork}

The heterogeneity in real-world data has led to growing interest in architectures that support representation and learning in heterogeneous graphs~\cite{wang2022survey, phan2022heterogeneous}. Various approaches address this heterogeneity from different perspectives: \cite{phan2022heterogeneous} explores the integration of natural language and code snippets, focusing on structural and semantic heterogeneity across mixed domains, while \cite{wang2022survey} models recommendation systems using heterogeneous graphs defined by multiple nodes and relation types. In~\cite{yang2021interpretable}, heterogeneity is examined as a combination of node and edge types within information networks, employing meta paths to interpret relational dependencies.

A comprehensive review of the state-of-the-art reveals that, while these studies provide valuable insights into managing diverse data types within heterogeneous graphs, most approaches are limited to independently estimating graphs for each data type—categorical, sequential, or numerical, among others~\cite{zhou2020graph, phan2022heterogeneous}. 
However, while some GNN frameworks support multiple data types, there is still no unified architecture for the combined integration of continuous and discrete data in a single model designed for classification or prediction tasks~\cite{zhou2020graph, wang2022survey, yang2021interpretable}. 
Graph estimation across varied data contexts, as well as developing representations that integrate multiple heterogeneous data types, remains an open area, especially in MTS analysis. Despite recent advancements, challenges persist in learning from heterogeneous graphs that integrate continuous and discrete data. This integration, compounded by temporal irregularity, introduces distinct challenges for inference and learning in graph-based models.

Further exploration of state-of-the-art methods for managing ST relationships in graph-based architectures reveals two dominant approaches: spectral-based and spatial-based methods~\cite{zhou2020graph, sahili2023spatio}. ST-GNNs can also be categorized by how they incorporate temporal variation—either through an auxiliary ML model specifically for temporal dependencies or by embedding time directly within the graph structure~\cite{sahili2023spatio}. Hybrid ST-GNNs often combine spatial modules, such as spectral graph networks, spatial GNNs~\cite{ruiz2020gated}, or graph transformers, with temporal aspects captured by models like RNNs or transformers~\cite{sahili2023spatio}. Another approach embeds temporal information within the GNN itself, representing time as a signal, axis, subgraph, or through layer-stacking techniques~\cite{sahili2023spatio, liu2024todynet}. Despite recent advancements, a significant gap remains in developing models that fully integrate temporal dimensions within the graph structure, allowing GCNNs to process time as an intrinsic part of graph topology and thereby simplifying architectural complexity~\cite{zhou2020graph}.

After conducting a state-of-the-art review from a methodological perspective, in the clinical domain, traditional ML and Deep Learning (DL) models, such as Neural Networks (NNs), Gated Recurrent Units (GRUs), and transformers, have been widely used to address MDR prediction or simplifications thereof~\cite{martinez2022interpretable, escudero2024explainable}. 
While these models often achieve high accuracy in predicting MDR, many approaches focus on short-term patterns or individual input instances within limited contexts, restricting their ability to capture the complex temporal dependencies crucial for comprehensive MDR prediction~\cite{wang2023deep, tharmakulasingam2023transamr, nigo2024deep}.
Our previous work on MDR prediction in ICU settings focused on feature selection across independent time points~\cite{escudero2020temporal}, improving interpretability yet constraining temporal dependency use~\cite{martinez2022interpretable, escudero2021use}. More recently, we implemented a GRU model with explainable artificial intelligence methods adapted for irregular MTS data~\cite{escudero2024explainable}, enhancing interpretability but still lacking full integration of spatio-temporal relationships.

In contrast, recent graph-based approaches have shown significant promise in capturing complex clinical interactions, particularly for AMR and MDR prediction. These methods leverage GNNs to model intricate relationships among clinical, microbiological, and environmental factors, enhancing prediction accuracy for MDR cases~\cite{pi2022mdgnn, gouareb2023detection}. For instance, GNNs applied to predict MDR infections in Enterobacteriaceae have demonstrated advantages over traditional models~\cite{gouareb2023detection}, while GCNNs have enhanced performance in antiviral drug prediction~\cite{pi2022mdgnn}. However, most studies remain focused on specific organisms or resistance mechanisms rather than providing a broader framework for MDR prediction across multiple pathogens with heterogeneous and irregular MTS data. These limitations underscore the need for models incorporating dynamic temporal dependencies essential in diverse clinical scenarios.

Building on these advancements, our proposed XST-GCNN architecture uniquely integrates ST graph analysis to capture both temporal dynamics and spatial relationships within clinical data.his approach is aligned with recent developments in the medical field, such as~\cite{fu2023spatial}, which introduced an ST antibiogram predictor, and~\cite{senthilkumar2022dynamic}, which applied an ST-GNN for various healthcare applications.

\subsection{Contributions and Paper Outline}
\label{subsection:contrib&PaperOut}

This section outlines our main contributions beyond state-of-the-art, first methodologically and then in relation to MDR prediction. It also provides a roadmap for the architecture description and validation in the following sections.  

From the point of view of methodology, we introduce XST-GCNN, a graph-based DL architecture for irregular and heterogeneous MTS, with the following features:

\begin{itemize}[leftmargin=*]
    
    \item \textbf{Joint modeling of temporal and feature dependencies:} We propose an ST architecture that captures temporal and feature interactions within a unified architecture. The GCNN operates on a graph whose nodes are time-feature pairs, enhancing the representation of complex dependencies often overlooked in traditional methods.
    
    \item \textbf{Innovative graph estimation and GCNN design:} We explore several graph estimation techniques, including correlation-based methods, graph smoothness, and our novel Heterogeneous Gower Distance (HGD), designed for datasets with discrete and real-valued variables and compatible with Dynamic Time Warping (DTW). These approaches are used to construct two types of graph representations: i) a Cartesian Product Graph (CPG), which preserves stable feature relationships over time, and ii) an ST Graph (STG), which adapts the feature-to-feature relationship for each time step.   
    \item \textbf{Adaptive GCNN architecture for ST data:} At each layer, our GCNN, which supports both CPG and STG representations, can implement either a simple aggregation by processing the data with the normalized adjacency matrix or a more sophisticated higher-order polynomial filter able to deal with heterophilic data. This adaptability allows the model to generalize across diverse datasets, balancing complexity, accuracy, and robustness.
        
    \item \textbf{Emphasis on explainability alongside accuracy:} In addition to high predictive accuracy, we prioritize interpretability. Our model identifies key feature-time step combinations, providing insights into classification outcomes. Interpretability stems from the model's transparent design, while explainability uses supplementary techniques to clarify feature impact. Together, these qualities aid clinical decision-making and contribute to building trust in AI-driven predictions.

\end{itemize}

From the point of view of applicability, we validate the architecture in the specific context of MDR prediction using ST data from EHRs, demonstrating its effectiveness in real-world clinical scenarios. Relevant contributions in this regard include:

\begin{itemize}[leftmargin=*]
    \item \textbf{Clinical decision support.} XST-GCNN enhances clinical decision-making by identifying feature-time step combinations that are essential for accurate MDR predictions, thereby improving model interpretability and providing actionable insights for clinicians.

    \item \textbf{Superior predictive performance.} XST-GCNN outperforms traditional ML and DL approaches in classifying MDR in ICU patients, demonstrating its effectiveness in improving clinical outcomes.

    \item \textbf{Innovative application in MDR prediction.} This work represents a novel approach in healthcare analytics by applying graph-based methodologies specifically tailored for MDR prediction, highlighting the flexibility and robustness of the XST-GCNN architecture.
\end{itemize}

The remainder of this paper is organized as follows. Section~\ref{sec:Methods} details the methods and architectures within the XST-GCNN architecture, Section~\ref{sec:Results&Discussion} describes the case study and experimental validation, and Section~\ref{sec:Conclusion&Futurework} concludes with key findings and future research directions.

\section{Proposed Data Processing Architecture}
\label{sec:Methods}

\begin{figure*}[ht]
    \centering
	\centering
	\includegraphics[width=0.75\textwidth]{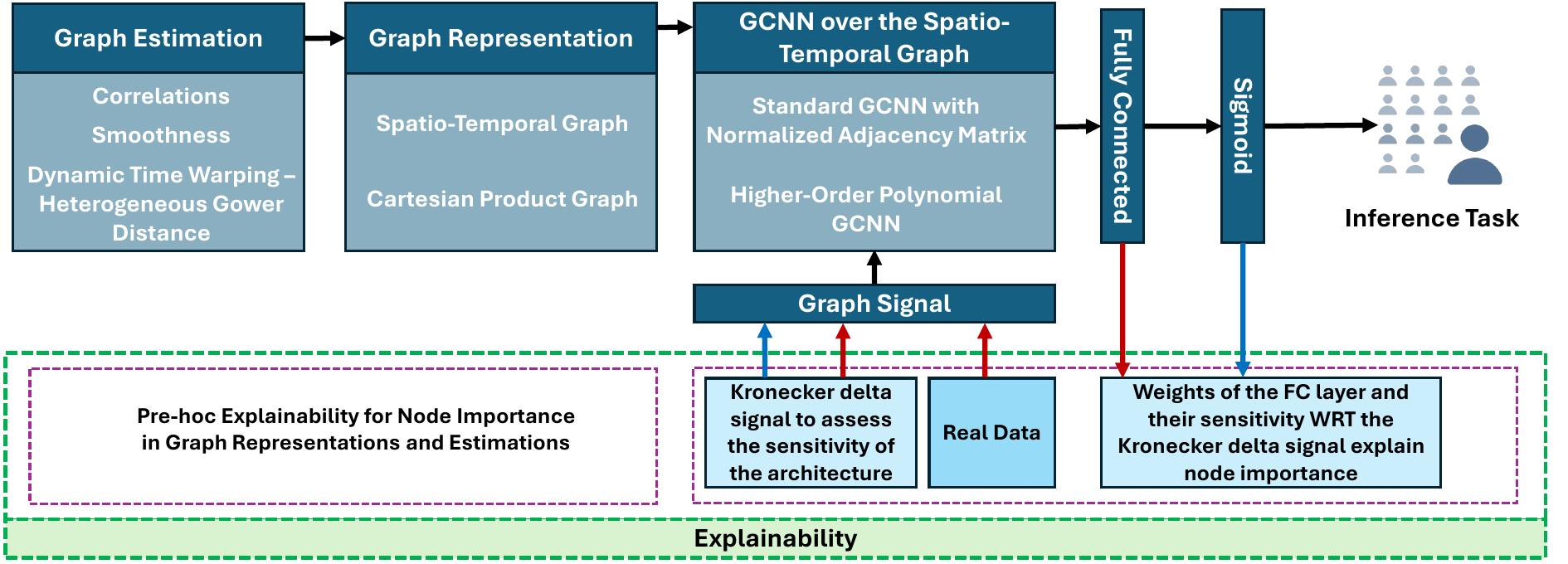}
    \caption{Proposed XST-GCNN architecture for inference tasks using irregular MTS and heterogeneous features. The architecture employs relatively advanced SP techniques, including graph estimation based on correlations, smoothness constraints, and distance measures such as HGD and DTW. The graph representation is modeled as an STG or CPG, capturing both temporal and spatial dependencies. Two definitions for the GCNN layer are proposed: Standard GCNNs with Normalized Adjacency and Higher-Order Polynomial GCNNs. These layers are followed by LeakyReLU activation and dropout layers before passing through Fully Connected (FC) layers and a sigmoid activation for the final inference task. The architecture also emphasizes explainability, incorporating both pre-hoc and intrinsic methods. Pre-hoc explainability is achieved through node importance analysis during the graph representation and estimation phases. 
    Intrinsic explainability is provided through analysis on both real and synthetic data during and after the architecture training. This includes the consideration of synthetic Kronecker delta signals to assess the sensitivity of the architecture with respect to each of the inputs. The combined approach put forth contributes to improved decision-making and a deeper understanding of the model's behavior.}
	\label{fig:pipeline}
\end{figure*}

This section introduces the proposed XST-GCNN architecture, specifically designed to address the challenges of irregular MTS and heterogeneous features, with a focus on EHR data. As shown in Fig.~\ref{fig:pipeline}, the architecture is meticulously crafted to capture these complexities while enhancing interpretability. We begin by defining essential notation, followed by an in-depth discussion of the graph learning and representation techniques employed. The section concludes with two specific GCNN designs tailored to process irregular MTS, yielding predictions that exploit the ST dependencies inherent in the dataset used.

\subsection{Notation}
\label{notation}

We define our patient dataset as $\mathcal{D} = \{(\mathbf{X}_p, {y}_{p})\}_{p=1}^P$, where $P$ denotes the total number of patients. The $p$-th patient is characterized by a feature matrix $\mathbf{X}_p \in \mathbb{R}^{F \times T}$, with $F$ representing the number of features and $T$ the number of time steps. 
The $t$-th column of $\mathbf{X}_p$, denoted by $[\mathbf{X}_p]_{(:,t)}$, is a vector comprising the $F$ features of patient $p$ at time $t$. Conversely, the $f$-th row of $\mathbf{X}_p$, denoted by $[\mathbf{X}_p]_{(f,:)}$, represents the TS of feature $f$ for the $p$-th patient across all $T$ time steps. 
Notice that in most clinical applications, some of the features are real-valued while others are binary. 
This presents several challenges, including the selection of proper metrics to construct graphs, which will be discussed in more detail in subsequent sections. Moving on to the labels, patients are classified based on the presence of MDR pathogens during their ICU stay. Specifically, patients with at least one positive culture for MDR are assigned to the MDR class, while those without are classified as non-MDR. In this binary classification task, $y_p = 1$ indicates patient $p$ developed MDR, and $y_p = 0$ indicates a non-MDR patient. The model's task is to predict these labels, with the predicted soft label for the $p$-th patient being denoted as $\hat{y}_p \in [0,1]$. 

The input MTS considered in this paper is heterogeneous and two-dimensional, which challenges traditional DL approaches. As explained in detail next, our approach is to build an ST graph that jointly models dependencies across features and time steps, and processes the information with a GCNN that leverages convolutions in the ST graph to integrate the information collected in the input MTS. 

\subsection{Graph-Learning Methods}
\label{sub:GLM}
Having established the notation, we now delve into the fundamental concepts of (directed) weighted graphs and the methods employed to estimate these graphs, which are pivotal to our architecture. A weighted graph is represented as $\mathcal{G} = (\mathcal{V}, \mathcal{E}, \mathcal{W})$, where $\mathcal{V} = \{1, \ldots, N\}$ denotes the set of nodes, $\mathcal{E} \subseteq \mathcal{V} \times \mathcal{V}$ represents the set of edges, and $\mathcal{W}: \mathcal{E} \to \mathbb{R}_+$ is a weight function assigning positive real values to each edge~\cite{isufi2024graph}. These weights indicate the strength or capacity of the connections between nodes. 

Graphs can be categorized as either directed or undirected. In a directed graph, edges have a specific orientation, denoted by pairs $(i, j) \in \mathcal{E}$, indicating a connection from node $i$ to node $j$~\cite{west2001introduction}.
Directed graphs distinguish between out-neighbors and in-neighbors, represented as $\mathcal{N}_i^{\text{out}} = \{j \in \mathcal{V} \mid (i, j) \in \mathcal{E}\}$ and $\mathcal{N}_i^{\text{in}} = \{j \in \mathcal{V} \mid (j, i) \in \mathcal{E}\}$, respectively. Conversely, an undirected graph is a special case where each pair of nodes is connected in both directions, represented by a symmetric adjacency matrix~\cite{diestel2024graph}. In undirected graphs, the neighboring set of a node $i$ is defined as $\mathcal{N}_i = \{j \in \mathcal{V} \mid (i, j) \in \mathcal{E}\}$. 
The graph is commonly represented by a weighted adjacency matrix \(\mathbf{A}\), an \(N \times N\) matrix where \([\mathbf{A}]_{ij}  > 0\) indicates the weight of the edge \((i, j) \in \mathcal{E}\)~\cite{isufi2024graph}. This matrix may be asymmetric for directed graphs, while it remains symmetric for undirected graphs. In symmetric graphs, another popular matrix is the graph Laplacian $\bbL$, which is defined as $\bbL=\bbD -\bbA$, where $\bbD$ is the (diagonal) degree matrix that satisfies 
\(\ \mathbf{D} = \text{diag} (\mathbf{A}\mathbf{1})\). 

Within this paper the estimated graphs are weighted and, when considering time dependencies, directed. 
Two graph representation approaches are considered. In the first one, each node represents a particular feature-time tuple \((f,t)\), so that \(\mathcal{V} = \{1, \ldots, F\} \times \{1,\ldots,T\}\). In the second one, each node represents a particular feature \(f\), so that \(\mathcal{V} = \{1, \ldots, F\}\). Depending on the graph representation approach, graph estimation can be performed by either focusing on the information from a specific time step, denoted as \(\mathbf{X'}_t \in \mathbb{R}^{F \times P}\), or considering the aggregated information from all patients over time, represented by the tensor \(\mathcal{X'} \in \mathbb{R}^{F \times T \times P}\). 

With these considerations in mind, the next step is to discuss the methods employed to estimate the weights associated with the edges of these graphs (see Fig.~\ref{fig:pipeline}). The proposed methods for graph computation include: a) correlation-based methods, b) graph smoothness-based methods, and c) HGD-DTW. Thus, the initial step involves a formal introduction to each of these methodologies, with adaptations specific to the MDR context. We differentiate between two distinct approaches for graph estimation: i) analyzing the entire temporal horizon as a unified entity, and ii) examining each temporal step as an independent unit. These methodologies facilitate the assessment of feature relationships by utilizing both aggregated and time-specific data, thereby enhancing the understanding of dynamic interactions within the dataset.

\subsubsection{Correlation-Based Methods} 
A simple yet effective method to draw links between pairs of nodes is to quantify the level of the correlation between the features associated with the nodes. Since we consider a heterogeneous setting where some of the features are binary and some are real-valued, we implement three distinct methods to capture the level of association: a) the Pearson correlation coefficient~\cite{cohen2009pearson}, used when both features are real-valued; b) the Matthews correlation coefficient (aka Phi coefficient ~\cite{malik2013family}), used when both features are binary; and c) the Point-Biserial correlation coefficient~\cite{malik2013family}, used when one of the variables is binary and the other is real-valued. Next, we briefly review each of these three methods. In the following, we assume that we focus on nodes $i=1$ and $j=2$, with \(\mathbf{z}_1 = [z_1^{(1)}, z_1^{(2)}, \ldots, z_1^{(K)}] \in \mathbb{R}^{1 \times K}\) and \(\mathbf{z}_2 = [z_2^{(1)}, z_2^{(2)}, \ldots, z_2^{(K)}] \in \mathbb{R}^{1 \times K}\) denoting the two generic signals associated with those two nodes, and \(K\) representing a generic vector length. 

\begin{itemize}[leftmargin=*]
\item The \textit{Pearson correlation coefficient}~\cite{cohen2009pearson} quantifies the linear relationship between two numerical (real-valued) features \(\mathbf{z}_1\) and \(\mathbf{z}_2\). With \(\overline{z}_1\) and \(\overline{z}_2\) representing their respective means over the \(K\) observations, the normalized Pearson correlation coefficient is simply given by
\begin{equation}\label{NPC}
r_{pc}(\mathbf{z}_1, \mathbf{z}_2) = \frac{\sum_{k=1}^{K} (z_1^{(k)} - \overline{z}_1) (z_2^{(k)} - \overline{z}_2),
}{\sqrt{\sum_{k=1}^{K} (z_1^{(k)} - \overline{z}_1)^2 \sum_{k=1}^{K} (z_2^{(k)} - \overline{z}_2)^2}}.
\end{equation}

\item The \textit{Phi coefficient} assesses the level of association between two \textit{binary} features \(\mathbf{z}_1\) and \(\mathbf{z}_2\) $\in \mathbb{R}^{1 \times K}$~\cite{malik2013family}. Let \(n_{ij}(\bbz_1,\bbz_2)\) be the frequency of observations corresponding to each binary state combination of \(\mathbf{z}_1\) and \(\mathbf{z}_2\). Specifically, \(n_{11}(\bbz_1,\bbz_2)\) and \(n_{00}(\bbz_1,\bbz_2)\) indicate the counts where both vectors simultaneously take the values ``1'' and ``0'', respectively, while \(n_{10}(\bbz_1,\bbz_2)\) and \(n_{01}(\bbz_1,\bbz_2)\) capture the instances of mixed states. Furthermore, let \(n_{1}(\bbz_1)\) and \(n_{0}(\bbz_1)\) denote the total counts where the input vector (in this case \(\mathbf{z}_1\)) is ``1'' or ``0''. Then, the Phi coefficient of the pair $(\mathbf{z}_1, \mathbf{z}_2)$ is 
\begin{equation}
r_{\phi}(\mathbf{z}_1, \mathbf{z}_2) \!=\! \frac{n_{11}(\bbz_1,\bbz_2) n_{00}(\bbz_1,\bbz_2) - n_{10}(\bbz_1,\bbz_2) n_{01}(\bbz_1,\bbz_2)}{\sqrt{\!n_{1}(\bbz_1) n_{0}(\bbz_1) n_{1}(\bbz_2) n_{0}(\bbz_2)}}.
\label{Phi}
\end{equation}
The numerator in \eqref{Phi} reflects the difference in the joint occurrences of concordant states, while the denominator normalizes this difference by the product of the total occurrences, ensuring a scale-invariant measure of association.

\item Finally, the \textit{Point-Biserial coefficient}~\cite{malik2013family} is used when one feature (say $\bbz_1$) is real-valued and the other one (say $\bbz_2$) is binary. Let \(s_{z_1}\) be the standard deviation of the numerical feature \(\mathbf{z}_1\), \(\overline{z}_1^1(\bbz_1,\bbz_2)\) the mean of feature \(\mathbf{z}_1\) for the entries where \(\mathbf{z}_2\) is ``1", and \(\overline{z}_1^0(\bbz_1,\bbz_2)\) its counterpart for the entries where \(\mathbf{z}_2\) is ``0". Moreover, as in \eqref{Phi}, the terms \(n_{1}(\bbz_2)\) and \( n_{0}(\bbz_2)\) denote the number of ``1's" and ``0's" in \(\mathbf{z}_2\), respectively. Then, the Point-Biserial coefficient of the pair $(\mathbf{z}_1, \mathbf{z}_2)$ is 
\begin{equation}
r_{pb}(\mathbf{z}_1, \mathbf{z}_2)\!=\! \frac{\overline{z}_1^1(\bbz_1,\bbz_2) - \overline{z}_1^0(\bbz_1,\bbz_2)}{s_{z_1}} \sqrt{\!\frac{ n_{1}(\bbz_2)  n_{1}(\bbz_2)}{( n_{1}(\bbz_2)+ n_{0}(\bbz_2))^2}}.
\label{PB}
\end{equation}
\end{itemize}

The next step is to use the coefficients in \eqref{NPC}, \eqref{Phi}, and \eqref{PB} to build the graph. We consider first the approach where a different graph is learned for every $t$. To that end, we focus on $\bbX_t'\in\reals^{F\times P}$, which is one slice of tensor $\ccalX'$. The first step is to remove (mask) the columns of $\bbX_t'$ associated with patients that, for time step $t$, do not have information, giving rise to the matrix $\bbX_{\text{masked},t} = \mathbf{X''} \in \reals^{F\times P_t}$ with $P_t\leq P$. Since the rows of \(\mathbf{X'}_{\text{masked},t}\) represent features, we compute the adjacency of the feature-to-feature graph $\bbA_t$ by: a) computing an $F\times F$ matrix $\bbW_t$ whose entry $(f,f')$ is obtained using\footnote{The particular choice will depend on the nature of the $(f,f')$ pair, using \eqref{NPC} if both are real-valued, \eqref{Phi} is both are binary, and \eqref{PB} if they are mixed.}\label{f:footnotecoefficients} \eqref{NPC}-\eqref{PB} and b) setting to zero all the entries $[\bbW_t]_{ij}$ such that $|[\bbW_t]_{ij}|\leq\eta_t$, with $\eta_t$ being a pre-specified threshold. Finally, this procedure is repeated for $t=1,...,T$ giving rise to the set of graphs with adjacency matrices $\{\bbA_t\}_{t=1}^T$. The procedure for the case 
where a single graph is used to represent the full dataset is quite similar. To that end, we first rearrange the data in tensor \(\mathcal{X'} \in \mathbb{R}^{F \times T \times P}\) into the matrix \(\mathbf{X''} \in \mathbb{R}^{F \times TP}\). Then, we remove (mask) the columns of $\mathbf{X''}$ associated with $(t,p)$ pairs with missing information, giving rise to the matrix $\mathbf{X''}_{\text{masked}} = \mathbf{X''} \in \reals^{F\times K}$ with $K$ representing here the number of columns with data once the missing information was removed. After this, we create a single $F\times F$ matrix $\bbW$, so that the $(i,j)$-th entry of  $\bbW$ is set to the Pearson/Phi/Point-biserial coefficient between the $i$-th and $j$-th rows of $\mathbf{X''}_{\text{masked}}$ (see footnote \ref{f:footnotecoefficients}). Finally, the entries of $\bbW$ whose magnitude is below a threshold are set to zero, giving rise to the \textit{static} adjacency matrix $\bbA$.

\subsubsection{Smoothness-Based Graph Estimation} Smoothness-based graph estimation aims to learn graphs where the given signals are smooth~\cite{dong2014learning,kalofolias2016learn,chepuri2017learning}. While different ways to measure signal variability exist, the most popular in the graph SP literature is $\bbx^T\bbL\bbx=\sum_{i,j}[\bbA]_{ij}(x_i-x_j)^2$~\cite{ortega2018graph}. When a set of $K$ graph signals is given, the latter can be written as $\sum_{k=1}^K\bbx_k^T\bbL\bbx_k=\tr(\sum_{k=1}^K\bbx_k\bbx_k^T\bbL)=K\tr(\hat{\bbC}\bbL)$, with $\hat{\bbC}$ denoting the sample covariance matrix. To ensure that all the features contribute the same, the sample covariance is typically normalized, so that smoothness-based estimation aims at learning a graph that minimizes $\tr(\hat{\bbC}_{\text{norm}}\bbL)$,  where \([\hat{\bbC}_{\text{norm}}]_{ij} =[\hat{\bbC}]_{ij}/\sqrt{([\hat{\bbC}]_{ii}[\hat{\bbC}]_{jj}}\). 

Among the different smoothness-based graph learning methods we adopt the one in~\cite{chepuri2017learning}, which implements a greedy approach to learn the edges of the graph. More specifically, the method starts with an FC graph (i.e., a graph with $F(F-1)/2$ edges), and removes the edge that reduces the graph signal variability the most. The process is repeated iteratively until a pre-specified value of edges (or smoothness) is reached. 

As we discussed after~\eqref{PB}, the next step is estimating the graph in two different setups. In the first one, the goal is to learn a graph for every time $t$. To that end, we use as graph signals the columns of $\mathbf{X'}_{\text{masked},t} \in \mathbb{R}^{F \times P_t}$, form the sample covariance matrix  $\hat{\bbC}_{\text{norm},t}$, and then use that matrix to learn $\bbA_t$ via~\cite{chepuri2017learning}. In the second one,  the goal is to learn a single graph $\bbA$. The graph signals in this case are the columns of matrix $\mathbf{X''}_{\text{masked}} = \mathbf{X''} \in \reals^{F\times K}$, which are used to learn the single matrix $\hat{\bbC}_{\text{norm}}$. 

\subsubsection{Heterogeneous Gower Distance - Dynamic Time Warping} 
Another popular way to build graphs is using a distance function so that two nodes are connected if the distance between the signals (features) associated with those nodes is below a given threshold. Taking into account the particularities of our data, here we implement a distance-based graph learning method where we: i) use the HGD, which is an adaptation of the Gower distance~\cite{podani1999extending} we propose and explain below, to measure the distance between heterogeneous features, and ii) when measuring distances between TS (i.e., when learning a single static graph), we combine HGD with DTW, a technique used to measure the dissimilarity between temporal sequences~\cite{muller2007dynamic, escudero2023dtwparallel}. A key feature in DTW is that the sequences may be misaligned, which is often the case in applications such as speech or EHR data.

First, let us consider two generic one-dimensional vectors $\bbz_1\in\reals^K$ and $\bbz_2\in\reals^K$. To enhance consistency in the comparison of heterogeneous variables, we propose specific modifications to the HGD.  When both  \( \bbz_1 \) and \( \bbz_2 \) are continuous variables, normalization is achieved by defining the maximum values  \(z_1^{\max}=\max\{\{z_1^{(k)}\}_{k=1}^K\} \), \( z_2^{\max}=\max\{\{z_2^{(k)}\}_{k=1}^K\} \) and \(z_{1,2}^{\max}=\max\{z_1^{\max},z_2^{\max}\}\), and then, with a slight abuse of notation, 
rescaling each variable as \( z_1^{(k)} = z_1^{(k)}\frac{  z_{1,2}^{\max}}{z_{1}^{\max}} \) and \( z_2^{(k)} = z_2^{(k)} \frac{  z_{1,2}^{\max}}{z_{2}^{\max}} \) for all $k$. In cases where one variable  (say $\bbz_1$) is binary, and the other one (say $\bbz_2$) is continuous, the binary variable is rescaled by setting $z_1^{(k)}=\max\{\{z_2^{(k)}\}_{k=1}^K\}$ if $z_1^{(k)}=1$ and $z_1^{(k)}=\min\{\{z_2^{(k)}\}_{k=1}^K\}$ if $z_1^{(k)}=0$, ensuring that both variables lie within the same range.
Then, the HGD between those vectors is defined as 
\begin{equation}
\delta_{\text{HGD}}(\bbz_1, \bbz_2) = \frac{1}{K} \sum_{k=1}^{K} \frac{|z_1^{(k)} - z_2^{(k)}|}{R_k},
\label{gw}
\end{equation}
where $R_k$ is the dynamic range of the $k$-th entry of the vector among all the vectors in the dataset~\cite{podani1999extending}. Using this definition, each of the graphs in $\{\bbA_t\}_{t=1}^T$ is learned by computing an $F\times F$ matrix $\check{\bbW}_t$ whose $(i,j)$-th entry is $[\check{\bbW}_t]_{ij}=\delta_{\text{HGD}}([\bbX_t']_{(i,:)},[\bbX_t']_{(j,:)})$. After this, we apply an exponential transformation, so that the weights are found as
\begin{equation}\label{e:exponential_transformation_edge_weights}
    [\bbW_t]_{ij}=e^{-\beta[\check{\bbW}_t]_{ij}^2},
\end{equation}
with $\beta$ being a temperature parameter used to tune the sensitivity of the graph with respect to the distance. 
Clearly, since the transformation in \eqref{e:exponential_transformation_edge_weights} is monotonically decreasing, smaller distances give rise to higher weights. Finally, a thresholding operator is applied entry-wise to set to zero edges with a small weight (i.e., nodes that are far apart from each other).

We move now to the setup where we use DTW to learn a single (static) graph. Specifically, let $\check{\bbX}_f\in\reals^{P\times T}$ be the slice of tensor $\ccalX'$ that contains the values of feature $f$ for all patients and time steps. Clearly, $\check{\bbX}_f$ can be understood as an MTS with $P$ values per time step. The goal here is to use DTW to learn the feature-to-feature graph $\bbA$. Specifically, we build an $F\times F$ distance matrix $\check{\bbW}$ whose $(i,j)$-th entry is $[\check{\bbW}]_{ij}=DTW_{\text{HGD}}(\check{\bbX}_i,\check{\bbX}_j)$, with $DTW_{\text{HGD}}$ representing the DTW distance computed using the HGD. To explain how this distance is computed, let us define first the cumulative distance matrix $\mathbf{M} \in \mathbb{R}^{(T+1) \times (T+1)}$, whose values are obtained using the following initialization and recursive procedure~\cite{seto2015multivariate}:
\begin{align}
    [\mathbf{M}]_{1,1} &= 0,~[\mathbf{M}]_{1,t+1} = \infty,~[\mathbf{M}]_{t+1,1} = \infty,~\forall~t \label{eq:Matrix_DTW_step1} \\
    [\mathbf{M}]_{t,t'} &= \delta_{\text{HGD}}([\check{\bbX}_i]_{(:,t)}, [\check{\bbX}_j]_{(:,t')}) \notag \\
    &\quad + \min\{[\mathbf{M}]_{t-1,t'-1}, [\mathbf{M}]_{t-1,t'}, [\mathbf{M}]_{t,t'-1}\}.\label{eq:Matrix_DTW_step2}
\end{align}
After filling $\mathbf{M}$ column by column (or row by row), the DTW distance is obtained as $DTW_{\text{HGD}}(\check{\bbX}_i,\check{\bbX}_j) = [\mathbf{M}]_{T+1, T+1}$; see, e.g.,~\cite{seto2015multivariate} for additional details and motivation. While most implementations of DTW consider scalar TS and Euclidean distances, the distance in each step can be adapted for the dataset at hand  (in this case HGD). The main advantage of DTW over correlation and smoothness metrics in previous distances is that DTW is able to deal with misaligned data.

\subsection{Graph-Representation Approaches}
\label{GRA}
The graph-learning methods detailed in the previous section captured relationships between features. The goal in this section is to explain how to leverage those results to deal with graphs able to capture both spatial (i.e., feature-to-feature) and temporal dynamics. The ultimate goal is to develop tractable graph-based representations that effectively capture the intrinsic relationships within irregular MTS data representation (see Fig.~\ref{fig:pipeline}). To that end, we consider two distinct approaches: one that leverages the time varying graphs $\{\bbA_t\}_{t=1}^T$ (labeled as STG), and another one that leverages the static graph $\bbA$ (labeled as CPG).

\noindent\textbf{Spatio-Temporal Graph (STG):} Our goal here is to describe a graph $\ccalG_{STG}$ whose nodes represent $(f,t)$ tuples and, as a result, is represented by the $FT\times FT$ adjacency matrix $\bbA_{STG}$. The first $F$  columns (rows) index the features associated with the first time step, the second $F$ columns (rows) the features associated with the second time step, and so forth. For the STG approach, we consider that: i) the relation among features changes over time, and the strength of this relationship is given by $\{\bbA_t\}_{t=1}^T$, and ii) the value of any feature at time step $t$ is related to the value of the same feature at the previous time step $t-1$. This results in the following adjacency matrix
\begin{equation}\label{e::adjacency_STG}
\mathbf{A}_{STG} = \begin{bmatrix}
\mathbf{A}_1 & \mathbf{I} & \mathbf{0} & \cdots & \mathbf{0} \\
\mathbf{0} & \mathbf{A}_2 & \mathbf{I} & \ddots & \vdots \\
\vdots & \ddots & \ddots & \ddots & \mathbf{0} \\
\vdots & \ddots & \ddots & \ddots & \mathbf{I} \\
\mathbf{0} & \cdots & \cdots & \mathbf{0}  & \mathbf{A}_T
\end{bmatrix},
\end{equation}
where \(\mathbf{I}\) denotes the $F\times F$ identity matrix that establishes temporal connections between features at consecutive time steps. Notice that the model in \eqref{e::adjacency_STG} is directed, since $\bbA_{STG}\neq \bbA_{STG}^\top$. If additional information on the time evolution of the MTS exists, e.g., by assuming that the MTS can be modeled as an autoregressive (AR) process of other one, matrix $\bbI$ can be replaced with the AR transition matrix.

\noindent\textbf{Cartesian Product Graph (CPG):} A related representation approach can be implemented using the static feature-to-feature $\bbA$ graph as input. As in the previous case, the goal is to build an $FT\times FT$ adjacency matrix, labeled in this case as $\bbA_{CPG}$. For the CPG approach, we consider that: i) the relation among features does not change over time and the strength of this relationship is given by $\bbA$, and ii) the value of any feature at time step $t$ is related to the value of the same feature at the previous time step. This results in a matrix $\bbA_{CPG}$ that is obtained by replacing $\bbA_t$ with $\bbA$ for all $t=1,...,T$ in \eqref{e::adjacency_STG}. 

Interestingly, this construction is equivalent to saying that the graph $\ccalG_{STG}$ is obtained by computing the CPG between the static feature-to-feature graph and the \emph{directed} path graph of $\ccalG_{dp}$ of length $T$ \cite{sandryhaila2014big}. To be more specific, $\ccalG_{dp}$ is a graph with $T$ nodes whose adjacency matrix is given by $[\bbA_{dp}]_{t,t+1}=1$ for $t=1,...,T-1$ and zero for all other entries. Clearly, $\ccalG_{dp}$ is directed, has only $T-1$ edges, and encodes the temporal progression. Using standard results of graph-theory, if two graphs are combined using the CPG, the resulting adjacency matrix can be obtained as
\begin{equation}\label{e::adjacency_CPG}
\mathbf{A}_{CPG} = \mathbf{A}_{dp} \oplus \mathbf{A}, \end{equation}
where \(\oplus\) represents the Kronecker sum. One advantage of the ST structure in \eqref{e::adjacency_CPG} is that the \textit{spectral} properties of $\mathbf{A}_{CPG}$ follow directly from those of $\bbA_{dp}$ and $\bbA$ \cite{sandryhaila2014big}, facilitating the analysis and processing of signals defined over $\mathbf{A}_{CPG}$. 

The graph-representations introduced in this section can be leveraged to model the data matrix $\bbX_p$ (i.e., the information associated with patient $p$) as a graph signal defined over either $\ccalG_{STG}$ or $\ccalG_{CPG}$. Both approaches integrate temporal dynamics within the spatial graph structure. The selection between $\bbA_{STG}$ and $\bbA_{CPG}$ will depend on the specificities of the application. The STG approach is particularly advantageous when: i) the relations between features are complex and vary significantly over time and ii) the number of samples (patients) for each time step is sufficiently high so that the time-varying graphs can be effectively estimated. In contrast, the CPG approach is more suitable when: i) the relations between features do not change too much over time, ii) data is limited, and iii) graph spectral tools are important to process the data at hand.

\subsection{Graph Convolutional Neural Network}
\label{GCNN}

Upon constructing the ST graphs, the next step is to develop graph-based processing and learning architectures capable of incorporating both spatial and temporal dependencies. 
Considering the heterogeneity found in our input data—numerical and binary variables—as well as the success of NNs models in MDR prediction~\cite{martinez2022interpretable, escudero2024explainable}, the strategy proposed in this paper is to create two GCNN architectures that take advantage of the ST graphs described in~\eqref{e::adjacency_STG} and \eqref{e::adjacency_CPG}.

The numerical experiments in Section~\ref{sec:Results&Discussion} will showcase that, by leveraging the ST relationships in the learned graphs, the architectures proposed next enhance significantly predictive accuracy and interpretability, thereby facilitating more informed decision-making in clinical settings. 

Succinctly, the goal of our architectures is to predict the output $\hat{y}_p$ for the input $F\times T$ matrix $\bbX_p$, which is the data associated with patient $p$. To that end, we first vectorize $\bbX_p$ and, then, use $\text{vec}(\bbX_p)$ as an input for a GCNN operating over a graph with $FT$ nodes [cf. \eqref{e::adjacency_STG} and \eqref{e::adjacency_CPG}]. Finally, we apply an FC layer to transform the output of the GCNN into the estimated label $\hat{y}_p$. Numerous GCNNs have been proposed in the literature \cite{wu2020comprehensive}, two of which are considered in this paper. The first one is the classical GCNN proposed in \cite{kipf2016semi}, which at every layer only considers linear averaging among one-hop neighbors. In contrast, the second architecture, at every layer, implements a polynomial graph filter with learnable coefficients \cite{gama2020graphs,ioannidis2020tensor}, enabling the mixing of information from multiple-hop neighborhoods and learning of low-pass/band-pass/high-pass frequency responses tailored the properties of the dataset at hand. The remaining of this section is organized as follows. We first explain the two GCNNs considered, along with their main differences. Then, we describe our full DL architecture, which incorporates the previous GCNNs as the key processing block.

\vspace{.1cm}
\noindent\textbf{GCNN-1: Standard GCNN with Normalized Adjacency Matrix.} This formulation leverages an architecture based on graph convolutional layers utilizing the normalized adjacency matrix. The core layer of this model normalizes the adjacency matrix \( \mathbf{A} \) by adding self-loops and incorporating the degree diagonal matrix \( \hat{\mathbf{D}} = \text{diag} ((\mathbf{A} + \mathbf{I})\mathbf{1}) \), leading to 
\begin{equation}\label{e::adjacencyNormalized_and_withselfloops}
\hat{\mathbf{A}} = \hat{\mathbf{D}}^{-\frac{1}{2}} (\mathbf{A}+\mathbf{I}) \hat{\mathbf{D}}^{-\frac{1}{2}},
\end{equation}
where \(\mathbf{I}\) is the identity matrix. This normalization is critical for stabilizing the training process and enhancing the effectiveness of information propagation, as evidenced in prior studies on GCNNs. The graph convolution operation at each layer is expressed as
\begin{equation}\label{e::layerGNN_KipfArchitecture}
\mathbf{H}^{(l+1)} = \hat{\mathbf{A}} \mathbf{H}^{(l)} \mathbf{W}^{(l)},    
\end{equation}
where \( \mathbf{H}^{(l)}\in\reals^{FT\times U_l} \) denotes the data matrix at layer \( l \), and \( \mathbf{W}^{(l)}\in \reals^{U_l\times U_{l+1}} \) is the trainable weight matrix. The number of rows in $\mathbf{H}^{(l+1)}$ coincides with the number of nodes ($FT$, for the setup at hand). In contrast, the number of columns in $\mathbf{H}^{(l+1)}$ represents the number of ``synthetic'' graph signals generated by layer $l+1$. With this in mind, the formalization of this architecture is given by
\begin{subequations}\label{ee:layersGCNN1}
\begin{align}
&\mathbf{H}^{(1)} = \sigma(\hat{\mathbf{A}} \mathbf{X} \mathbf{W}^{(0)} + \mathbf{1}\mathbf{b}^{(0)}),\\
&\mathbf{H}^{(l+1)} = \sigma(\hat{\mathbf{A}} \mathbf{H}^{(l)} \mathbf{W}^{(l)} + \mathbf{1}\mathbf{b}^{(l)}), \quad \text{for } l = 1, \ldots, L-1,
\end{align}
\end{subequations}
where \( \mathbf{X} \in \reals^{FT\times U_0} \) represents the $U_0$ input graph signals (in our case, $U_0=1$);  \( \sigma \) is a nonlinear scalar activation function applied entry-wise; $\mathbf{1}$ is a column vector of all ones; and \( \mathbf{b}^{(l)}\in\reals^{1\times U_{l+1}} \) is the learnable bias vector.  Overall, the learnable parameters are $\{\bbW^{(l)}\in \reals^{U_l\times U_{l+1}} \}_{l=0}^{L-1}$ and $\{\bbb^{(l)}\in \reals^{1\times U_{l+1}} \}_{l=0}^{L-1}$. 

\vspace{.1cm}
\noindent\textbf{GCNN-2: Higher-Order Polynomial GCNN.} While effective in many relevant applications, the architecture in \eqref{ee:layersGCNN1} suffers from problems associated with oversmoothing and poor performance dealing with heterophilic datasets~\cite{yan2022two}. Motivated by this, we propose a second GCNN architecture, which, at every layer, implements (a bank of) polynomial graph filters with learnable coefficients \cite{gama2020graphs,ioannidis2020tensor}. This formulation extends the standard GCNN by enabling: i) higher-order graph convolutions that linearly mix information from nodes that are multiple hops away and ii) learning generic frequency responses, which mitigates the problems associated with oversmoothing and endows the GCNN to be applied to non-homophilic datasets. Both generalizations open the door to a GCNN able to capture more complex relationships within the graph. The higher-order convolution operation is defined as \cite{gama2020graphs}
\begin{equation}\label{ee:polynomialgraphconvolution}
\mathbf{H}^{(l+1)} = \sum_{k=0}^{K-1} \mathbf{A}^k \mathbf{H}^{(l)} \mathbf{W}_k^{(l)},
\end{equation}
with $\mathbf{A}^k$ denoting the $k$-th power of $\bbA$. 
In contrast with \eqref{e::layerGNN_KipfArchitecture}, here we apply the adjacency matrix multiple times, which is an effective way to mix information within a $(K-1)$-hop neighborhood. Additionally, the number of learnable weights per convolution is $K\times U_l \times U_{l+1}$, endowing the architecture with additional degrees of freedom that can be used to learn more general graph-based transformations. 

Upon replacing \eqref{ee:polynomialgraphconvolution} into a GCNN with $L$ layers, we have that
\begin{subequations}\label{ee:layersGCNN2}
\begin{align}
&\mathbf{H}^{(1)} = \sigma\left( \sum_{k=0}^{K-1} \hat{\mathbf{A}}^k \mathbf{X} \mathbf{W}_k^{(0)} + \mathbf{1}\mathbf{b}^{(0)} \right),\\
&\mathbf{H}^{(l+1)} \!= \!\sigma\left( \sum_{k=0}^{K-1} \hat{\mathbf{A}}^k \mathbf{H}^{(l)} \mathbf{W}_k^{(l)} + \mathbf{1}\mathbf{b}^{(l)} \right), \; \text{for } l \!=\! 1,..., L-1,
\end{align}
\end{subequations} 
where \( \mathbf{X}\in\reals^{FT\times U_0} \) are the $U_0$ input graph signals; \( \sigma \) denotes the nonlinear entry-wise activation function; and $\mathbf{b}^{(l)}\in\reals^{1\times U_{l+1}}$ is the bias vector. Since the weight matrix is different for each $k$, this GCNN is able to assign positive or negative weights for the information of $1,...,K-1$ neighbors. This contrasts with \eqref{e::layerGNN_KipfArchitecture}, which always assigns a positive weight to the information of 1-hop neighbors. In a nutshell, \eqref{ee:layersGCNN2} enables the model to learn and leverage more sophisticated graph structures, thereby enhancing its capacity to model complex relationships.

The two alternative GCNN definitions presented here provide a way to adapt the ST DL model to the particularities and complexities of the data at hand. Each definition serves different scenarios, with GCNN-1 providing a more straightforward approach suitable for general tasks, and GCNN-2 offering a more powerful framework for capturing intricate graph-based dependencies.

\vspace{.1cm}
\noindent\textbf{Proposed ST graph-based DL architecture.} As already pointed out, our goal is to design a DL architecture for binary classification leveraging an ST graph whose links capture the strength of the relation between feature-time pairs. The (MTS) input to the architecture is the $F\times T$ matrix $\bbX_p$ and the (soft label) output  is the scalar $\hat{y}_p\in [0,1]$. The architecture (cf. Fig.~\ref{fig:pipeline}) is composed of three blocks, which are applied sequentially. 
\begin{itemize}[leftmargin=*]
\item The first block simply vectorizes the input and replaces missing values with zero, giving rise to the $FT$-dimensional vector $\bbx_p^{\mathrm{zp}}=\mathrm{zeropadd}(\mathrm{vec}(\bbX_p))$. 
\item  The second block implements one of the two GCNNs presented in this section, with $L$ denoting the number of layers. The GCNN architecture is specified as follows. 
\begin{enumerate}
    \item[a)] 
No pooling is implemented and, as a result, the output of all the GCNN layers (including the last one) can be interpreted as $FT$-dimensional signals defined over the ST graph. 
    \item[b)]  The input of the first layer is the graph signal $\bbx_p^{\mathrm{zp}}\in\reals^{FT\times 1}$; the layers $l=1,...,L-1$ generate as output multiple graph signals, with the number of generated signals per layer $U_l$ being set to a constant $U$ whose value is considered a hyperparameter; and the $L$-th layer outputs a single signal (i.e., we set $U_L=1$), so that the output of the GCNN $\bbh^{(L)}=\bbH^{(L)}\in\reals^{FT\times 1}$ is a one-dimensional graph signal. For the GCNN-1 architecture, the learnable parameters are the bias vectors $\{\bbb^{(l)}\in\reals^{1\times U_l}\}_{l=0}^{L-1}$ along with the weight matrices $\{\bbW^{(l)}\in\reals^{U_l\times U_{l+1}}\}_{l=0}^{L-1}$, with $U_0=U_L=1$ and $U_l=U$ otherwise. For the GCNN-2 architecture, the learnable parameters are the bias vectors $\{\bbb^{(l)}\in\reals^{1\times U_l}\}_{l=0}^{L-1}$ along with the weights $\{\bbW_k^{(l)}\in\reals^{U_l\times U_{l+1}}\}_{l=0}^{L-1}$ for $k=0,...,K-1$.
    \item[c)]  The activation function applied at each layer is set to a LeakyReLU, which is defined as \(\text{LeakyReLU}(h) = h \) if \( h > 0\), and \(\alpha h \) if \( h \leq 0\), where \( \alpha \) is a small positive scalar, typically set to \( \alpha = 0.01 \). This activation function is particularly suitable for GCNN-1, since it has been shown to mitigate the vanishing gradient problem by maintaining a non-zero gradient when \( h \) is negative, thereby enhancing gradient flow during backpropagation.
    \item[d)]  After applying the activation function, a dropout mechanism is introduced. Dropout operates by randomly deactivating a fraction \( \pi \) of the features for each node during training, effectively performing model averaging and preventing co-adaptation of feature representations. Mathematically, the operation performed by the Dropout layer can be expressed as \( [\bbH_{\mathrm{input}}^{(l+1)}]_{i,u} = [\bbH^{(l)}]_{i,u} [\bbR_\pi]_{i,u} \), where \( \bbH^{(l)} \) represents the output of the \( l \)-th layer, \( \bbH_{\mathrm{input}}^{(l+1)} \) is the input to the \( (l+1) \)-th layer, and \( \bbR_\pi\) is a random binary matrix where each entry is independently drawn from a Bernoulli distribution with parameter \( \pi \). Here, \( [\bbR_\pi]_{i,u} = 0 \) indicates that the feature \( u \) of node \( i \) is dropped out. Note that if no dropout is applied, we simply have \( \bbH_{\mathrm{input}}^{(l+1)} = \bbH^{(l)} \), as considered in \eqref{ee:layersGCNN1} and \eqref{ee:layersGCNN2}.
\end{enumerate}
\item The third block implements an FC layer. Specifically, if $\bbh^{(L)}\in\reals^{FT}$ denotes the output of the last layer of the GCNN, this block estimates the soft output as
\begin{equation}
\hat{y}_p = \sigma(\mathbf{w}_o^\top \bbh^{(L)} + b_o) = \frac{1}{1 + e^{-(\mathbf{w}_o^\top \mathbf{h}^{(L)} + b_o)}},
\label{eq:sigmoid_fc}
\end{equation}
with $\bbw_o\in\reals^{FT}$ and $b_o$ being learnable parameters, and the activation function corresponding to a sigmoid (binary softmax). The main reason to consider a simple FC layer is to foster the explainability of the architecture. Clearly, the entries of $\bbw_o$ indicate the relative importance that the information gathered in each of the entries of $\bbh^{(L)}$ has for the final classification. Specifically, more positive and larger weights indicate that the corresponding feature-time pair is more relevant to the MDR class. Conversely, more negative and larger weights suggest a strong association with the non-MDR class, decreasing the likelihood of the input being classified as MDR.  This formulation not only facilitates the final binary decision but also provides a clear interpretation of how each feature-time pair contributes to the classification outcome.
\end{itemize}

The next section assesses the accuracy performance of our ST graph-based architectures in a real-world dataset. As explained in detail next, our novel integration of ST dynamics into a GCNN sets a new benchmark for predictive modeling in dynamic environments, particularly in the context of healthcare applications.

\section{Results and Discussion}
\label{sec:Results&Discussion}

In this section, we apply the XST-GCNN architecture to a real-world dataset to classify MDR patients in the ICU setting at UHF. We first describe the dataset and outline the experimental setup, including parameter optimization. Next, we analyze graph properties and pre-hoc explainability, followed by an evaluation of XST-GCNN’s classification performance against state-of-the-art methods. Finally, we explore the model’s explainability, demonstrating how XST-GCNN clarifies the impact of each feature-time pair on classification outcomes, providing insights that support informed clinical decision-making.
\footnote{Due to space limitations, only a summary of the numerical results is included in the main manuscript. Additional results,  including figures, tables, and graph representations, are available in the supplementary material submitted together with the main manuscript. Furthermore, we also provide additional results, analysis, and the code for the complete set of experiments in our GitHub repository {\url{https://github.com/oscarescuderoarnanz/XST-GCNN}}.} 

\subsection{Dataset}
\label{subSec:Dataset}

This clinical case study uses the XST-GCNN architecture to predict MDR in ICU patients at UHF, aiming to detect the first MDR-positive culture within a 14-day window. The dataset spans 17 years, from January 2004 to February 2020, and includes the longitudinal clinical records of 3,502 ICU patients. Among these, 548 patients had at least one MDR-positive culture during their stay, highlighting a significant class imbalance.

A rigorous anonymization protocol was implemented to ensure patient confidentiality, with ethical approval obtained from the UHF Research Ethics Committee (ref: 24/22, EC2091). Building on this foundation, the primary objective is to solve a binary classification problem by predicting, based on data available within the first $T=14$ days, whether a patient will develop MDR. Due to variations in ICU stays—since not all patients remain hospitalized for the same number of days, nor do they develop MDR on the same day—MTS data exhibit irregularities that must be addressed in the analysis. The analysis is confined to ICU stays, excluding pre-admission data to focus on the transmission dynamics of MDR pathogens within the ICU environment.

To achieve this, microbiological cultures and antibiograms were conducted to identify MDR pathogens, with particular attention to the first MDR-positive culture detected. Patients without MDR were labeled as \(y_p = 0\), while those with a positive MDR culture within the first 14 days were labeled as \(y_p = 1\). The 14-day window was chosen for its clinical relevance, aligning with standard infection control practices where the risk of transmission and the application of treatment protocols are most critical. 

The dataset's richness is underscored by the extensive set of $F=80$ variables collected for each patient, which are crucial for understanding the factors contributing to MDR development and the overall ICU environment. These variables are organized into three main categories, providing a comprehensive foundation for the analysis:
\begin{itemize}[leftmargin=*]
    \item \textbf{Patient-specific antibiotic therapy}: To monitor daily antibiotic therapy in the ICU for each patient, binary variables were created to indicate whether the patient received specific antibiotic families. These families include Aminoglycosides (AMG), Antifungals (ATF), Intestinal anti-infectives (ATI), Antimalarials (ATP), Carbapenems (CAR), 1st, 2nd, 3rd, and 4th generation Cephalosporins (CF1, CF2, CF3, CF4), Glycyclines (GCC), Glycopeptides (GLI), Lincosamides (LIN), Lipopeptides (LIP), Macrolides (MAC), Monobactams (MON), Nitroimidazoles (NTI), unclassified antibiotics (\textit{Others}), Oxazolidinones (OXA), Miscellaneous (OTR), Broad-spectrum Penicillins (PAP), Penicillins (PEN), Polypeptides (POL), Quinolones (QUI), Sulfonamides (SUL), and Tetracyclines (TTC). The variable \textit{Others} denotes the administration of other antibiotics not included in this list.

    \item \textbf{ICU occupancy and co-patient treatments}: This group of variables captures essential environmental factors that reflect the overall health conditions and treatment protocols within the ICU. ``Co-patients" are defined as the other patients sharing the ICU with the $p$-th patient during the same time interval, excluding the patient under study. The variables include continuous data on the number of co-patients receiving each of the 25 antibiotic families, represented as ``\( \text{family}_{n} \)". Additionally, daily ICU occupancy is documented through three main variables: i) the total number of co-patients (\# of pat\textsubscript{tot}), ii) the number of co-patients diagnosed with MDR bacteria (\# of pat\textsubscript{MDR}), and iii) the number of co-patients undergoing any form of antibiotic therapy (\# of pat\textsubscript{atb}). These variables offer a detailed view of the ICU environment, providing insights into the overall health status and treatment practices within the unit.

    \item \textbf{Patient health monitoring variables}: This category includes both continuous and binary variables that serve as key indicators of patient health. Continuous variables monitor daily hours of mechanical ventilation, tracheostomy duration, ulcer presence, hemodialysis hours, and the number and types of catheters used, including Peripherally Inserted Central Catheters (PICC), Central Venous Catheters (CVC), and specific insertion sites such as right (R), left (L), subclavian (S), femoral (F), and jugular (J). Additionally, we include the Nine Equivalents of Nursing Manpower Use Score (NEMS), a patient severity scale utilized by nursing staff.
    Binary variables capture whether the patient received insulin, artificial nutrition, sedation, muscle relaxation, or underwent postural changes. Additionally, organ failures are closely monitored to identify specific dysfunctions—hepatic, renal, coagulation, hemodynamic, and respiratory—with the administration of vasoactive drugs also being recorded. These variables offer critical insights into the patient’s health status and the necessary interventions during their ICU stay. 
\end{itemize}

\subsection{Experimental Setup and Parameter Optimization}
\label{subSec:experimentalSetup}

The experimental setup was designed to evaluate the predictive performance of the XST-GCNN architecture. The dataset was divided into a training set ($\mathcal{D}_{\text{train}}$) with 70\% of the patients and a test set ($\mathcal{D}_{\text{test}}$) with the remaining 30\%. Given the class imbalance—where MDR-positive cases were underrepresented—an undersampling approach was used within $\mathcal{D}_{\text{train}}$ to balance the class distribution and reduce potential bias~\cite{he2009learning}. This method was chosen to maintain the integrity of the original data while minimizing the risk of overfitting. Combined with 5-fold cross-validation, this strategy enhanced the model’s generalization and reduced overfitting.

Hyperparameter optimization focused on fine-tuning the XST-GCNN architecture to maximize accuracy. We explored key hyperparameters, including dropout rates \{0.0, 0.15, 0.3\}, learning rates \{1e-3, 1e-2, 5e-2, 0.1\}, and learning rate decay \{0, 1e-5, 1e-4, 1e-3, 1e-2\}. The number of units in the hidden layers ranged from \{4, 8, 16, 32, 64\}, and the network depth varied between 1 and 6 layers to find the most effective configuration. For the GCNN component within \mbox{XST-GCNN}, which uses polynomial filter banks, we evaluated the polynomial order \(K\) by testing values of 2 and 3, as these represent the definitions of the GCNN we assessed.

Model performance was evaluated based on three key metrics—sensitivity, specificity, and Receiver Operating Characteristic - Area Under the Curve (ROC-AUC)~\cite{bradley1997use}. Sensitivity measured the model’s ability to correctly identify MDR cases, while specificity assessed its accuracy in detecting non-MDR cases. The ROC-AUC metric comprehensively evaluated the model's capability to distinguish between MDR and non-MDR cases. The results presented in Section~\ref{sec:Exp} were obtained using the test set ($\mathcal{D}_{\text{test}}$) and evaluated with all three metrics. To ensure the robustness and stability of the model's performance, each experiment was repeated three times with different random splits of the training and test sets, accounting for variability in data selection.

\subsection{Testing XST-GCNN in a Real-World Dataset}
\label{sec:Exp}
In this section, we assess some properties of the graph derived from our real dataset to understand feature interactions and data relationships. We then compare our approach with state-of-the-art methods and explore the explainability of the best-performing model using both synthetic and real signals\footnote{Additional results, figures, and tables can be found in the supplementary material and in the following folder of the GitHub repository {\url{https://github.com/oscarescuderoarnanz/XST-GCNN}}}.

\subsubsection{\textbf{Graph Properties and Pre-hoc Explainability}}
\label{subsec:graphProperties}

To understand in more detail the structure of the graphs generated by our XST-GCNN architecture, we analyzed two fundamental metrics in graph theory: edge density and edge entropy. Edge density, $\eta(\mathcal{G})$, quantifies graph connectivity by calculating the ratio of existing edges to the maximum possible number of edges~\cite{kolaczyk2014statistical}. For an undirected graph with $|\mathcal{V}|$ vertices and $|\mathcal{E}|$ edges, it is defined as $\eta(\mathcal{G}) = \frac{2|\mathcal{E}|}{|\mathcal{V}|(|\mathcal{V}|-1)}$, where a value of 1 indicates a complete graph and 0 indicates a graph without edges. Edge entropy, $H(\mathcal{G})$, measures the complexity of the graph's structure by assessing the distribution of edges, calculated as $H(\mathcal{G}) = -\sum_{i=1}^{|\mathcal{V}|} d_i \ln d_i$, where $d_i$ represents the normalized weighted degree of the $i$-th node~\cite{kolaczyk2014statistical}.

The analyses conducted, which evaluate edge density and edge entropy across various thresholds, demonstrate that selecting a threshold value of $0.975$ effectively ensures meaningful graph sparsity while preserving structural integrity. This threshold maintains low edge density and stable entropy, facilitating the identification of key relationships without excessive connectivity. By assessing multiple threshold values, we confirmed that the model preserves its performance and representational capacity with the chosen threshold, validating its robustness and utility in processing heterogeneous and irregular MTS. Additional information regarding the method's sensitivity to threshold selection is provided in the supplementary material (see Appendix A). 

Additionally, the resultant graphs were provided to clinical experts of the UHF, including the head of the ICU, who validated the relevance and accuracy of the ST graph representations. This implies that, even before training the architecture, the selected representation has the potential to help clinical experts to visually assess evolving patterns, highlighting connections between different features across time, and identifying which variables gain or lose relevance throughout the timeline. Further details on these representations and pre-hoc explainability can be found in Appendix B of our supplementary material.

\subsubsection{\textbf{Prediction Results}}
\label{PredResults}

The experimental analysis underscores the effectiveness of the proposed XST-GCNN architecture in classifying patients as MDR or non-MDR by leveraging ST relationships and heterogeneous features within EHR data. The results, presented in Table~\ref{tab:performance_comparison}, are benchmarked against baseline methods to assess the efficacy of various approaches rigorously. 

We start by analyzing the baseline models, which include the 
GRU~\cite{chung2014empirical}, Gated-Graph RNN (G-GRNN)~\cite{ruiz2020gated}, and GCNN approaches. The GRU demonstrated high specificity ($93.04 \pm 2.48$), indicating its effectiveness in correctly identifying non-MDR patients. However, its lower sensitivity ($58.49 \pm 4.08$) suggests a limitation in detecting true positives, leading to potential misclassification of MDR patients. The G-GRNN provided a more balanced performance with a ROC-AUC of $72.25 \pm 1.36$, sensitivity of $81.76 \pm 0.89$, and specificity of $62.74 \pm 3.44$ (Table~\ref{tab:performance_comparison}), reflecting better handling of ST dependencies. Among the GCNN approaches, the HGD-DTW criterion stood out, achieving the highest ROC-AUC of $74.53 \pm 0.94$ (Table~\ref{tab:performance_comparison}).

Next, we analyze the XST-GCNN architecture, starting with the models based on GCNN-1 (standard GCNN with normalized adjacency matrix). The CPG models show notable improvements under GCNN-1, particularly with the correlation criterion, achieving an ROC-AUC of $76.74 \pm 0.85$ (Table~\ref{tab:performance_comparison}). This suggests that leveraging feature correlations within the graph structure enhances the model's ability to distinguish between MDR and non-MDR cases. Additionally, the smoothness criterion improves sensitivity ($74.84 \pm 0.89$), indicating that smoother graph representations help in detecting MDR cases by reducing noise and highlighting consistent patterns.
The STG models within GCNN-1, particularly those using the HGD criterion, offer an ROC-AUC of $78.17 \pm 1.04$, further illustrating the potential of advanced graph-based criteria. However, these methods still underperform when compared to GCNN-2, which uses a higher-order polynomial approach. 

Moving to the models based on GCNN-2, there is a noticeable enhancement in performance. The CPG with correlation under GCNN-2 achieved an ROC-AUC of $80.59 \pm 4.79$ and specificity of $78.79 \pm 8.55$ (see Table~\ref{tab:performance_comparison}), effectively modeling complex spatial relationships. The STG approaches under GCNN-2, especially those using the HGD criterion, yielded the highest overall metrics with an ROC-AUC of $81.03 \pm 2.43$, sensitivity of $75.47 \pm 2.67$, and specificity of $78.11 \pm 0.95$ (see Table~\ref{tab:performance_comparison}). These results highlight the critical importance of capturing both ST relationships to significantly improve model accuracy.

In conclusion, the XST-GCNN architecture consistently outperforms traditional models such as GRU, G-GRNN, and other GCNN-based architectures. By integrating ST dynamics, XST-GCNN captures complex patterns missed by other approaches. The best results for both GCNN-1 and GCNN-2 are achieved with the HGD or HGD-DTW methods, highlighting HGD's effectiveness in handling heterogeneous and irregular MTS data. GCNN-2 generally surpasses GCNN-1, confirming its suitability for EHR data. Thus, XST-GCNN, combined with HGD, stands as a leading architecture for MDR patient classification, advancing beyond existing methods.

\begin{table}[ht]
\caption{Mean $\pm$ standard deviation of the performance (ROC-AUC, sensitivity, specificity, AUC) on 5 test sets when considering XST-GCNN architecture and baseline models for MDR versus non-MDR patient classification. The highest average performance for each figure of merit is in bold.}
    \centering
    \resizebox{\columnwidth}{!}{%
    \begin{tabular}{c|l|c|c|c}
        \toprule
        \multirow{2}{*}{\textbf{}} & \multirow{2}{*}{\textbf{Method}} & \multicolumn{3}{c}{\textbf{Performance Metrics}} \\
        \cmidrule(lr){3-5}
        &  & ROC-AUC & Sensitivity & Specificity \\
        \midrule 
        \multirow{7}{*}{\rotatebox[origin=c]{90}{Baselines}} 
        & GRU~\cite{chung2014empirical} &  80.78 $\pm$ 1.57 & 58.49 $\pm$ 4.08 & \textbf{93.04 $\pm$ 2.48} \\
        \cmidrule(lr){2-5}
        & G-GRNN~\cite{ruiz2020gated} &  72.25 $\pm$ 1.36 & \textbf{81.76 $\pm$ 0.89} & 62.74 $\pm$ 3.44 \\ 
        \cmidrule(lr){2-5}
        & GCNN-1 (correlations) &  69.91 $\pm$ 2.00 & 62.89 $\pm$ 3.56 & 66.44 $\pm$ 7.30 \\ 
        \cmidrule(lr){2-5}
        & GCNN-1 (smoothness) &  72.70 $\pm$ 1.87 & 63.52 $\pm$ 2.35 & 70.15 $\pm$ 1.11 \\ 
        \cmidrule(lr){2-5}
        & GCNN-1 (HGD-DTW) &  74.53 $\pm$ 0.94 & 61.01 $\pm$ 2.35 & 74.30 $\pm$ 0.42  \\ 
        \midrule 
        \multirow{18}{*}{\rotatebox[origin=c]{90}{XST-GCNN (our)}} & CPG with GCNN-1 (correlations) &  76.74 $\pm$ 0.85 & 72.33 $\pm$ 3.21 & 72.39 $\pm$ 1.67 \\
        \cmidrule(lr){2-5}
        & CPG with GCNN-1 (smoothness) &  76.80 $\pm$ 0.81 & 74.84 $\pm$ 0.89 & 73.63 $\pm$ 0.97 \\ 
        \cmidrule(lr){2-5}
        & CPG with GCNN-1 (HGD-DTW) &  77.77 $\pm$ 1.71 & 75.47 $\pm$ 2.67 & 72.73 $\pm$ 3.24 \\ 
        \cmidrule(lr){2-5}
        & STG with GCNN-1 (correlations) &  76.44 $\pm$ 1.01 & 74.84 $\pm$ 3.88 & 71.94 $\pm$ 3.55  \\ 
        \cmidrule(lr){2-5}
        & STG with GCNN-1 (smoothness) & 75.94 $\pm$ 1.20 & 72.33 $\pm$ 0.89 & 74.41 $\pm$ 0.99 \\ 
        \cmidrule(lr){2-5}
        & STG with GCNN-1 (HGD) & 78.17 $\pm$ 1.04 & 76.10 $\pm$ 3.88 & 72.28 $\pm$ 2.22 \\ 
        \cmidrule(lr){2-5}
        & CPG with GCNN-2 (correlations) & 80.59 $\pm$ 4.79 & 72.33 $\pm$ 4.71 & 78.79 $\pm$ 8.55 \\ 
        \cmidrule(lr){2-5}
        & CPG with GCNN-2 (smoothness) & 79.94 $\pm$ 1.67 & 69.18 $\pm$ 5.41 & 80.13 $\pm$ 4.59 \\ 
        \cmidrule(lr){2-5}
        & CPG with GCNN-2 (HGD-DTW) & 76.95 $\pm$ 1.53 & 72.33 $\pm$ 2.35 & 72.95 $\pm$ 1.11 \\ 
        \cmidrule(lr){2-5}
        & STG with GCNN-2 (correlations) & 80.25 $\pm$ 4.07 & 78.62 $\pm$ 3.21 & 74.30 $\pm$ 3.73 \\  
        \cmidrule(lr){2-5}
        & STG with GCNN-2 (smoothness) & 75.59 $\pm$ 2.88 & 71.70 $\pm$ 1.54 & 73.74 $\pm$ 2.35 \\ 
        \cmidrule(lr){2-5}
        & STG with GCNN-2 (HGD) & \textbf{81.03 $\pm$ 2.43} & 72.33 $\pm$ 2.35 & 78.68 $\pm$ 1.24 \\
        \bottomrule
    \end{tabular}%
    }
    \label{tab:performance_comparison}
\end{table}

\subsubsection{\textbf{Explainability}}

The XST-GCNN architecture achieves high accuracy in MDR patient classification, effectively managing the challenges posed by irregular MTS and heterogeneous features. Beyond its strong performance, the model offers enhanced explainability by illustrating the contribution of each feature-time pair to the classification results. We evaluated this explainability through a detailed analysis using both real patient data and synthetic signals, which allowed for a systematic assessment of the model’s robustness across different scenarios. The analysis focused on the model with the highest ROC-AUC (STG estimated with HGD and GCNN-2, as described in Section~\ref{PredResults}), identifying the key feature-time pairs that most significantly influence MDR classification. The primary findings are summarized here, while a more in-depth review---including additional models and data splits—can be found in the accompanying GitHub repository.

\begin{figure*}[ht]
        \centering
    	\includegraphics[width=1.6\columnwidth]{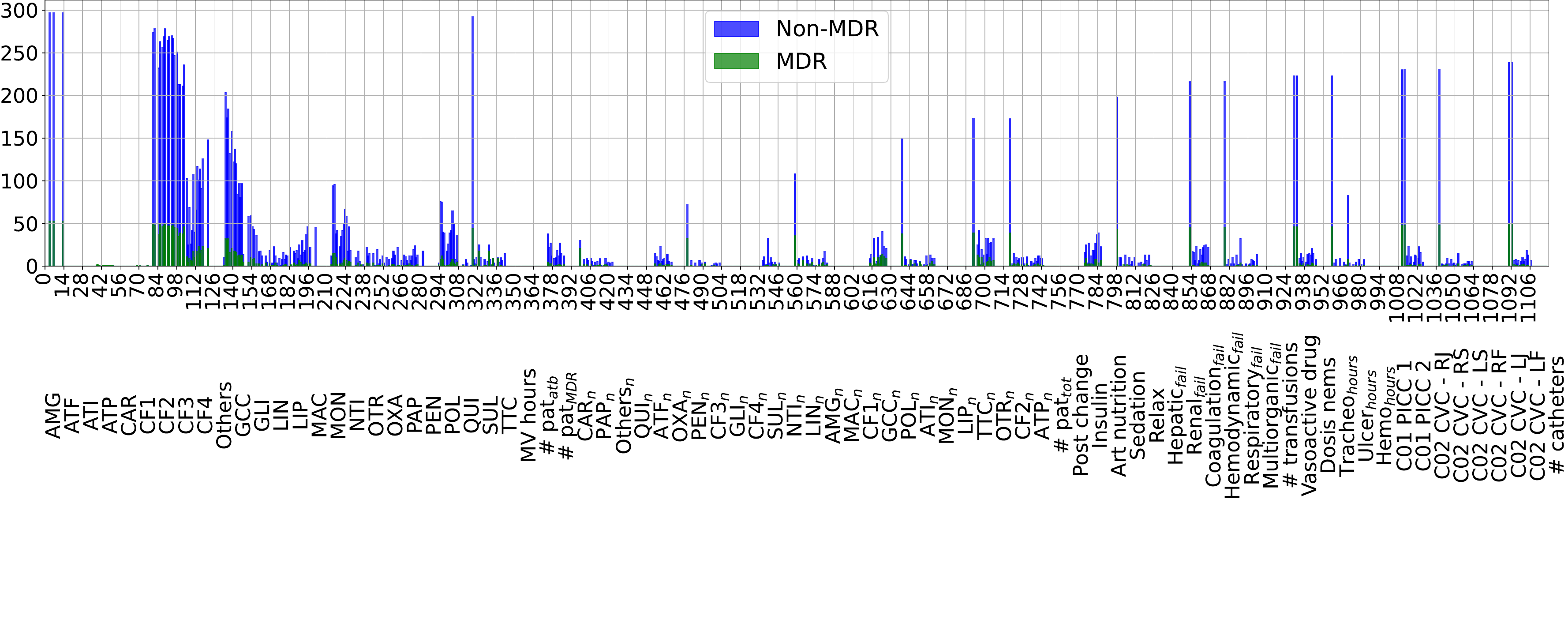}
        \vspace{-1cm}
        \caption{Bar graph depicts the frequency distribution of variables associated with MDR (green) and non-MDR (blue) cases across a range of clinical features. The x-axis represents different clinical variables, while the y-axis indicates the frequency of occurrence for each variable. This visual comparison highlights the prevalence and variation of specific clinical features between MDR and non-MDR groups, providing insights into potential risk factors and patterns associated with MDR.\vspace{-0.3cm}}
    	\label{fig:realSign}
    \end{figure*}

\paragraph{Analysis with Real Patient Data}\label{sec:realdata}
We start our analysis by trying to identify which $(f,t)$ pairs are more relevant for classifying patients as either MDR or non-MDR. To that end, we: i) use the absolute value of the product of the input and the weight of FC layer (cf. (\ref{eq:sigmoid_fc})) as the importance value and ii) deem as relevant the 56 pairs with highest importance value, which represents the 5\% of the $FT=1120$ pairs. In short, a) for each patient in the test set we compute the 1120 values at the input of the FC layer and select the top 56 $(f,t)$ pairs; and b) we then repeat the experiment for each test patient and count the number of times each $(f,t)$ pair is selected. The results are shown in Fig.~\ref{fig:realSign}. 
The x-axis represents the 1120 $(f,t)$ pairs, with the first 14 values being associated with the 14 measurements of feature AMG, the next 14 values with the 14 measurements of feature ATF, and so forth. The y-axis indicates the number of times each pair was selected in the top 5\%  across test samples, by class. High positive values strongly indicate MDR status, while negative weights correspond to non-MDR relevance, highlighting key variables influencing predictions and aiding clinical decision-making. The frequency distribution in Fig.~\ref{fig:realSign} reveals distinct patterns and risk factors differentiating these patient groups. Additional results and analysis of Fig.~\ref{fig:realSign} are available in the GitHub repository.\footnote{The complete analysis can be found at \url{https://github.com/oscarescuderoarnanz/XST-GCNN/tree/main/XST-GNN_Architecture/step3_GCNNs}, within each experiment’s folder.}

The main analysis of the results is as follows. For non-MDR patients, the most relevant variables were concentrated in the initial 4 time steps, particularly within the first 24 hours and again after the first 72 hours of admission. These include specific antibiotic treatments and the number of co-patients receiving the same antibiotics, such as POL, OTR, TTC, LIP, GCC, CF2, LIP, ATP, and ATI. Additionally, during these first 24 hours, variables related to the patient’s health status—such as hours on hemodialysis, tracheostomies, ulcers, and CO1 PICC 2, Co2 CVC - LJ and RF—emerged as significant. As time progressed from 24 to 72 hours, the total number of patients and those taking antibiotics, as well as NEMS, gained relevance, while after 72 hours post-admission, factors such as catheter types, patient status (NEMS), insulin, mechanical ventilation, respiratory failure, and the number of transfusions became increasingly important.

For patients with MDR infections, the 56 most relevant nodes were primarily identified within the initial 24 hours, with certain variables demonstrating importance across consecutive time points. Key variables in this critical period included the administration of antibiotic therapies and the number of co-patients receiving antibiotics from the same families, such as ATI, GCC, OTR, and TTC, with co-patient relevance observed for 21 out of 23 antibiotic families at the first time step, highlighting broader environmental impact. Additionally, patient health monitoring variables, including the number of catheters, CO2 CVC - RF, NEMS scores, hours on mechanical ventilation, and indications of hemodynamic, respiratory, and multi-organ failure, emerged as critical, correlating strongly with a more severe prognosis. The total number of MDR co-patients, overall ICU occupancy, and the total number of patients receiving antibiotics were also significant indicators of patient outcomes. Other important variables over time included co-patients receiving ATP and OTR in the last five time steps, CO2 CVC - LF from $t_4$ to $t_8$, and CO2 CVC - RF from $t_9$ to $t_{13}$, reflecting the increased complexity and severity commonly associated with MDR patients.

    
\paragraph{Analysis with Synthetic Signals}
We analyze the model's sensitivity by evaluating the activation response to individual input nodes (feature-time steps) in our pre-trained architecture. More specifically, as illustrated in Fig. \ref{fig:pipeline}, we generate 1120 Kronecker delta inputs and, for each of them, evaluate the impact in each of the entries of the output of the GNN $\bbh^{(L)}$ as well as in the decision made by the fully connected layer by computing the value of $\mathbf{w}_o^\top \mathbf{h}^{(L)}$ in \eqref{eq:sigmoid_fc}. This sensitivity analysis contributes to understanding the influence of each input value on the model’s predictions, verifying that the relationships learned from real data are faithfully represented in the model's behavior. 


The complete details and visualization of the results obtained can be accessed in our GitHub repository. Below, we summarize the main insights derived from analyzing the values $\mathbf{w}_o^\top \mathbf{h}^{(L)}$, focusing on three cases: (i) large values below zero, (ii) values above zero, and (iii) values close to zero.
In case (i), large positive weights show higher relevance within the first 48 hours for variables such as C02 CVC - RJ and RS, insulin, vasoactive drugs, and multi-organ failure. We have information on antibiotics taken by co-patients, specifically AMG and ATF, as well as whether a patient receives antibiotics like ATP, LIN, LIP, OXA, Others, PEN, and QUI. After the first 48 hours of patient admission, antibiotics such as ATP, CF1, CF3, MAC, MON, NTI, POL, and QUI begin to gain greater importance. Health monitoring variables in this case include C02 CVC - LF, coagulation failure, and hemodynamic, hepatic, respiratory, and multi-organ conditions. Additionally, insulin administration, postural changes, and vasoactive drugs remain relevant. In case (ii), large negative values initially highlight the first 48 hours, with notable relevance for information on antibiotics taken by co-patients, such as CAR, CF3, Others, PAP, and QUI, as well as the total number of patients and those receiving antibiotics. CF1 and OXA also contribute significantly. For health monitoring variables, transfusion count, hours with C02 CVC - LJ, and presence of postural changes emerge as key factors, alongside indicators of hemodynamic and hepatic failure. After 48 hours, the relevance shifts towards variables like antibiotics AMG, ATF, CAR, CF3, GLI, LIP, Others, PEN, QUI, and SUL concerning co-patients, among which CF2 is uniquely significant. Health status variables gain importance, including hours with C02 CVC - LF, and instances of coagulation, multi-organ, and respiratory failure, along with information on insulin, postural changes, relaxation, sedation, and vasoactive drugs.
Lastly, case (iii) encompasses values close to zero, corresponding to many node-time steps with minimal impact on class determination, indicating a stable or low-relevance state across most temporal intervals. For further details and a comprehensive visualization of these findings, please refer to our GitHub repository: \url{https://github.com/oscarescuderoarnanz/XST-GCNN/tree/main/XST-GNN_Architecture/step3_GCNNs}.

These analyses underscore the importance of managing specific interventions, especially those related to vascular access and antibiotic administration, to effectively manage patients at risk of developing MDR infections. These insights not only enhance predictive accuracy but also provide clinicians with a deeper understanding of the critical factors influencing patient outcomes, facilitating more informed and effective decision-making.

\paragraph{Clinical Relevance of Explainability}
The explainability provided by the XST-GCNN architecture is both technically robust and clinically significant. For non-MDR patients, emphasis on early indicators such as specific antibiotic treatments, the number of co-patients receiving the same antibiotics, and health status factors like catheter use and mechanical ventilation underscores the importance of preventive, early interventions. In contrast, the MDR class is characterized by variables reflecting antibiotic pressure, ICU occupancy, and the patient’s overall severity, including sedation, postural changes, and multi-organ failure. These distinct patterns not only differentiate patient groups but also offer clinicians critical insights for informed decision-making. Beyond identifying these patterns, our architecture clarifies the specific factors that drive classification outcomes for each patient, highlighting the variables relevant to MDR or non-MDR predictions. This explainability enables clinicians to anticipate MDR patient needs and tailor more aggressive, targeted treatments, thereby optimizing MDR management and ultimately improving patient outcomes.

\section{Conclusions and Future Work}
\label{sec:Conclusion&Futurework}

In this work, we proposed a novel graph-based DL architecture specifically designed to process irregular and heterogeneous MTS data. Our approach jointly modeled dependencies between features and time through an ST architecture, where a GCNN operates on a graph that integrates both temporal and feature dimensions. A primary contribution was our innovative use of HGD for graph estimation, effectively modeling the complexities of heterogeneous data by accurately representing both categorical and real-valued features. Additionally, we explored and compared various methods for defining the GCNN and estimating the graph structure, evaluating their impact on model performance. Beyond accuracy, we emphasized explainability by designing inherently interpretable architectures and conducting detailed analyses to illuminate the model’s decision-making process, facilitating more informed decisions.

We validated the XST-GCNN model through a real-world case study on predicting MDR in ICU patients, leveraging ST EHR data. Our model demonstrated a marked improvement over traditional state-of-the-art ML and DL models, showcasing predictive accuracy and practical relevance for healthcare analytics. Specifically, the XST-GCNN, which integrated STG estimated via HGD and a Higher-Order Polynomial GCNN, achieved $81.03 \pm 2.43$ (ROC-AUC), $72.33 \pm 2.35$ (sensitivity), and $78.68 \pm 1.24$ (specificity). These results outperform the best baseline model, a GRU, which attained $80.78 \pm 1.57$ (ROC-AUC), $58.49 \pm 4.08$ (sensitivity), and $93.04 \pm 2.48$ (specificity). A major limitation of the GRU was the significant imbalance between sensitivity and specificity, whereas our proposed architecture generated a more equitable distribution across these metrics. In addition to improving the binary decision-making process, the XST-GCNN provided interpretable insights into how specific feature-time pairs contribute to the classification, further enhancing its clinical applicability and transparency.

Further analysis of the estimated graphs and their clinical relevance confirmed that the architecture effectively captured critical patterns and variables essential for MDR prediction. The most relevant explainability findings obtained were related to the early administration of certain antibiotics, such as CAR, and the number of co-patients receiving similar treatments within the first 24 hours, which were highly predictive of MDR outcomes. Additionally, variables associated with organ failure, including decreased renal function and respiratory failure, were identified as key indicators. These patterns were consistently observed in both real-world ICU data and synthetic tests, highlighting the model’s ability to detect meaningful clinical signals that are essential for predicting MDR status.

Looking ahead, the future development of XST-GCNN will focus on several key areas. First, we aim to extend the applicability of the model to other domains where irregular MTS, heterogeneous data, and explainability are crucial. While the model was designed to be domain-agnostic, testing it in various fields beyond clinical data will help evaluate its robustness and effectiveness. In addition, we plan to incorporate more advanced explainability mechanisms, such as Graph Attention Networks\cite{velickovic2017graph}, which will enhance the model's ability to highlight the most important relationships in the data, improving explainability across both clinical and non-clinical settings. Another important area of focus will be optimizing the computational efficiency of XST-GCNN, making it suitable for deployment in resource-constrained environments. This will increase its accessibility to a wider range of healthcare institutions, particularly those with limited computational resources. Finally, we plan to explore combining the current architecture with RNNs by replacing the FC layer with an RNN. This modification will further enhance the model’s capacity for processing sequential data, allowing it to better capture temporal dependencies in more complex datasets.

In conclusion, XST-GCNN represents a significant advancement in the application of GNNs to clinical data, particularly for the prediction of MDR infections. By addressing the outlined challenges, including explainability and efficiency, and refining the approach, this research sets the stage for the development of more effective and reliable predictive models, with the potential to significantly impact patient care and clinical outcomes.

\bibliographystyle{IEEEtran}
\bibliography{references}

\begin{thebibliography}{10}
\providecommand{\url}[1]{#1}
\csname url@samestyle\endcsname
\providecommand{\newblock}{\relax}
\providecommand{\bibinfo}[2]{#2}
\providecommand{\BIBentrySTDinterwordspacing}{\spaceskip=0pt\relax}
\providecommand{\BIBentryALTinterwordstretchfactor}{4}
\providecommand{\BIBentryALTinterwordspacing}{\spaceskip=\fontdimen2\font plus
\BIBentryALTinterwordstretchfactor\fontdimen3\font minus \fontdimen4\font\relax}
\providecommand{\BIBforeignlanguage}[2]{{%
\expandafter\ifx\csname l@#1\endcsname\relax
\typeout{** WARNING: IEEEtran.bst: No hyphenation pattern has been}%
\typeout{** loaded for the language `#1'. Using the pattern for}%
\typeout{** the default language instead.}%
\else
\language=\csname l@#1\endcsname
\fi
#2}}
\providecommand{\BIBdecl}{\relax}
\BIBdecl

\bibitem{zhang2024graph}
X.~Zhang and Q.~Wang, ``A graph-assisted framework for multiple graph learning,'' \emph{IEEE Trans. Signal Inf. Process. Netw.}, 2024.

\bibitem{zhou2020graph}
J.~Zhou \emph{et~al.}, ``Graph neural networks: A review of methods and applications,'' \emph{AI open}, vol.~1, pp. 57--81, 2020.

\bibitem{wu2020comprehensive}
Z.~Wu \emph{et~al.}, ``A comprehensive survey on graph neural networks,'' \emph{IEEE Trans. Pattern Anal. Mach. Intell.}, vol.~32, no.~1, pp. 4--24, 2020.

\bibitem{zhang2021graph}
X.-M. Zhang, L.~Liang, L.~Liu, and M.-J. Tang, ``Graph neural networks and their current applications in bioinformatics,'' \emph{J. Biomed. Inform.}, vol.~12, p. 690049, 2021.

\bibitem{gilmer2017neural}
J.~Gilmer \emph{et~al.}, ``Neural message passing for quantum chemistry,'' \emph{arXiv preprint arXiv:1704.01212}, 2017.

\bibitem{isufi2021edgenets}
E.~Isufi, F.~Gama, and A.~Ribeiro, ``Edgenets: Edge varying graph neural networks,'' \emph{IEEE Trans. Pattern Anal. Mach. Intell.}, vol.~44, no.~11, pp. 7457--7473, 2021.

\bibitem{velickovic2017graph}
P.~Velickovic \emph{et~al.}, ``Graph attention networks,'' \emph{stat}, vol. 1050, no.~20, pp. 10--48\,550, 2017.

\bibitem{gama2020graphs}
F.~Gama, E.~Isufi, G.~Leus, and A.~Ribeiro, ``Graphs, convolutions, and neural networks: From graph filters to graph neural networks,'' \emph{IEEE Signal Process. Mag.}, vol.~37, no.~6, pp. 128--138, 2020.

\bibitem{yu2018spatio}
B.~Yu, H.~Yin, and Z.~Zhu, ``Spatio-temporal graph convolutional networks: A deep learning framework for traffic forecasting,'' \emph{arXiv preprint arXiv:1709.04875}, 2018.

\bibitem{ruiz2020gated}
L.~Ruiz, F.~Gama, and A.~Ribeiro, ``Gated graph recurrent neural networks,'' \emph{IEEE Trans. Signal Process.}, vol.~68, pp. 6303--6318, 2020.

\bibitem{AMR_deaths_2023}
K.~W.~K. Tang, B.~C. Millar, and J.~E. Moore, ``Antimicrobial resistance ({AMR}),'' \emph{Br. J. Biomed. Sci.}, vol.~80, 2023.

\bibitem{whoAntimicrobialResistance_2023}
{World Health Organization}, ``Antimicrobial resistance,'' \url{https://www.who.int/news-room/fact-sheets/detail/antimicrobial-resistance}, November 2023, [Accessed 27-05-2024].

\bibitem{world2024bacterial}
------, ``{WHO} bacterial priority pathogens list, 2024: {Bacterial} pathogens of public health importance to guide research, development and strategies to prevent and control antimicrobial resistance,'' \url{https://www.who.int/publications/i/item/9789240093461}, May 2024, [Accessed 27-05-2024].

\bibitem{shickel2017deep}
B.~Shickel, P.~J. Tighe, A.~Bihorac, and P.~Rashidi, ``Deep {EHR}: a survey of recent advances in deep learning techniques for electronic health record (ehr) analysis,'' \emph{IEEE J. Biomed. Health Inform.}, vol.~22, no.~5, pp. 1589--1604, 2017.

\bibitem{xie2022deep}
F.~Xie \emph{et~al.}, ``Deep learning for temporal data representation in electronic health records: A systematic review of challenges and methodologies,'' \emph{J. Biomed. Inform.}, vol. 126, p. 103980, 2022.

\bibitem{kallipolitis2023medical}
A.~Kallipolitis \emph{et~al.}, ``Medical knowledge extraction from graph-based modeling of electronic health records,'' in \emph{IFIP Int. Conf. Artif. Intell. Appl. Innov.}\hskip 1em plus 0.5em minus 0.4em\relax Springer, 2023, pp. 279--290.

\bibitem{murali2023towards}
L.~Murali, G.~Gopakumar, D.~M. Viswanathan, and P.~Nedungadi, ``Towards electronic health record-based medical knowledge graph construction, completion, and applications: A literature study,'' \emph{J. Biomed. Inform.}, vol. 143, p. 104403, 2023.

\bibitem{scarselli2008graph}
F.~Scarselli \emph{et~al.}, ``The graph neural network model,'' \emph{IEEE Trans. Pattern Anal. Mach. Intell.}, vol.~20, no.~1, pp. 61--80, 2008.

\bibitem{wang2022survey}
X.~Wang \emph{et~al.}, ``A survey on heterogeneous graph embedding: methods, techniques, applications and sources,'' \emph{IEEE Trans. Big Data}, vol.~9, no.~2, pp. 415--436, 2022.

\bibitem{phan2022heterogeneous}
H.~Phan and A.~Jannesari, ``Heterogeneous graph neural networks for software effort estimation,'' in \emph{Proc. 16th ACM/IEEE Int. Symp. Empir. Softw. Eng. Meas. (ESEM)}, 2022, pp. 103--113.

\bibitem{yang2021interpretable}
Y.~Yang \emph{et~al.}, ``Interpretable and efficient heterogeneous graph convolutional network,'' \emph{IEEE Trans. Knowl. Data Eng.}, vol.~35, no.~2, pp. 1637--1650, 2021.

\bibitem{sahili2023spatio}
Z.~A. Sahili and M.~Awad, ``Spatio-temporal graph neural networks: A survey,'' \emph{arXiv preprint arXiv:2301.10569}, 2023.

\bibitem{liu2024todynet}
H.~Liu \emph{et~al.}, ``Todynet: temporal dynamic graph neural network for multivariate time series classification,'' \emph{Inf. Sci.}, p. 120914, 2024.

\bibitem{martinez2022interpretable}
S.~Mart{\'\i}nez-Ag{\"u}ero \emph{et~al.}, ``Interpretable clinical time-series modeling with intelligent feature selection for early prediction of antimicrobial multidrug resistance,'' \emph{Future Gener. Comput. Syst.}, vol. 133, pp. 68--83, 2022.

\bibitem{escudero2024explainable}
{\'O}.~Escudero-Arnanz, C.~Soguero-Ruiz, J.~{\'A}lvarez-Rodr{\'\i}guez, and A.~G. Marques, ``Explainable artificial intelligence techniques for irregular temporal classification of multidrug resistance acquisition in intensive care unit patients,'' \emph{arXiv preprint arXiv:2407.17165}, 2024.

\bibitem{wang2023deep}
Y.~Wang \emph{et~al.}, ``A deep learning model for predicting multidrug-resistant organism infection in critically ill patients,'' \emph{J. Intensive Care}, vol.~11, no.~1, p.~49, 2023.

\bibitem{tharmakulasingam2023transamr}
M.~Tharmakulasingam \emph{et~al.}, ``Trans{AMR}: an interpretable transformer model for accurate prediction of antimicrobial resistance using antibiotic administration data,'' \emph{IEEE Access}, 2023.

\bibitem{nigo2024deep}
M.~Nigo \emph{et~al.}, ``Deep learning model for personalized prediction of positive mrsa culture using time-series electronic health records,'' \emph{Nat. Commun.}, vol.~15, no.~1, p. 2036, 2024.

\bibitem{escudero2020temporal}
{\'O}.~Escudero-Arnanz \emph{et~al.}, ``Temporal feature selection for characterizing antimicrobial multidrug resistance in the intensive care unit.'' in \emph{Eur. Conf. Artif. Intell.}, 2020, pp. 54--59.

\bibitem{escudero2021use}
------, ``On the use of time series kernel and dimensionality reduction to identify the acquisition of antimicrobial multidrug resistance in the intensive care unit,'' \emph{arXiv preprint arXiv:2107.10398}, 2021.

\bibitem{pi2022mdgnn}
J.~Pi, P.~Jiao, Y.~Zhang, and J.~Li, ``{MDGNN}: Microbial drug prediction based on heterogeneous multi-attention graph neural network,'' \emph{Front. Microbiol.}, vol.~13, p. 819046, 2022.

\bibitem{gouareb2023detection}
R.~Gouareb, A.~Bornet, D.~Proios, S.~G. Pereira, and D.~Teodoro, ``Detection of patients at risk of multidrug-resistant enterobacteriaceae infection using graph neural networks: A retrospective study,'' \emph{Health Data Sci.}, vol.~3, p. 0099, 2023.

\bibitem{fu2023spatial}
X.~Fu \emph{et~al.}, ``Spatial-temporal networks for antibiogram pattern prediction,'' in \emph{IEEE Int. Conf. Healthc. Inform. (ICHI)}.\hskip 1em plus 0.5em minus 0.4em\relax IEEE, 2023, pp. 225--234.

\bibitem{senthilkumar2022dynamic}
A.~Senthilkumar, M.~Gupte, and S.~Shridevi, ``Dynamic spatial-temporal graph model for disease prediction,'' \emph{Int. J. Adv. Comput. Sci. Appl.}, vol.~13, no.~6, 2022.

\bibitem{isufi2024graph}
E.~Isufi, F.~Gama, D.~I. Shuman, and S.~Segarra, ``Graph filters for signal processing and machine learning on graphs,'' \emph{IEEE Trans. Signal Process.}, 2024.

\bibitem{west2001introduction}
D.~B. West, \emph{Introduction to Graph Theory}.\hskip 1em plus 0.5em minus 0.4em\relax Prentice Hall, 2001.

\bibitem{diestel2024graph}
R.~Diestel, \emph{Graph Theory}.\hskip 1em plus 0.5em minus 0.4em\relax Springer (print ed.); Reinhard Diestel (eBooks), 2024.

\bibitem{cohen2009pearson}
I.~Cohen \emph{et~al.}, ``Pearson correlation coefficient,'' \emph{Noise Reduction Speech Process.}, pp. 1--4, 2009.

\bibitem{malik2013family}
S.~Malik and R.~Singh, ``A family of estimators of population mean using information on point bi-serial and phi correlation coefficient,'' \emph{arXiv preprint arXiv:1302.1658}, 2013.

\bibitem{dong2014learning}
X.~Dong, D.~Thanou, P.~Frossard, and P.~Vandergheynst, ``Learning laplacian matrix in smooth graph signal representations,'' \emph{IEEE Trans. Signal Process.}, vol.~62, no.~20, pp. 5271--5284, 2014.

\bibitem{kalofolias2016learn}
V.~Kalofolias, ``How to learn a graph from smooth signals,'' in \emph{Int. Conf. Artif. Intell. Statist. (AISTATS)}, 2016, pp. 920--929.

\bibitem{chepuri2017learning}
S.~P. Chepuri, S.~Liu, G.~Leus, and A.~O. Hero, ``Learning sparse graphs under smoothness prior,'' in \emph{IEEE Int. Conf. Acoust., Speech, Signal Process. (ICASSP)}.\hskip 1em plus 0.5em minus 0.4em\relax IEEE, 2017, pp. 6508--6512.

\bibitem{ortega2018graph}
A.~Ortega \emph{et~al.}, ``Graph signal processing: Overview, challenges, and applications,'' \emph{Proceedings of the IEEE}, vol. 106, no.~5, pp. 808--828, 2018.

\bibitem{podani1999extending}
J.~Podani, ``Extending gower's general coefficient of similarity to ordinal characters,'' \emph{Taxon}, vol.~48, no.~2, pp. 331--340, 1999.

\bibitem{muller2007dynamic}
M.~M{\"u}ller, ``Dynamic time warping,'' \emph{Inf. Retrieval Music Motion}, pp. 69--84, 2007.

\bibitem{escudero2023dtwparallel}
{\'O}.~Escudero-Arnanz \emph{et~al.}, ``dtwparallel: A python package to efficiently compute dynamic time warping between time series,'' \emph{SoftwareX}, vol.~22, p. 101364, 2023.

\bibitem{seto2015multivariate}
S.~Seto, W.~Zhang, and Y.~Zhou, ``Multivariate time series classification using dynamic time warping template selection for human activity recognition,'' in \emph{IEEE Symp. Ser. Comput. Intell.}, 2015, pp. 1399--1406.

\bibitem{sandryhaila2014big}
A.~Sandryhaila and J.~M. Moura, ``Big data analysis with signal processing on graphs: Representation and processing of massive data sets with irregular structure,'' \emph{IEEE Signal Process. Mag.}, vol.~31, no.~5, pp. 80--90, 2014.

\bibitem{kipf2016semi}
T.~N. Kipf and M.~Welling, ``Semi-supervised classification with graph convolutional networks,'' \emph{arXiv preprint arXiv:1609.02907}, 2016.

\bibitem{ioannidis2020tensor}
V.~N. Ioannidis, A.~G. Marques, and G.~B. Giannakis, ``Tensor graph convolutional networks for multi-relational and robust learning,'' \emph{IEEE Trans. Signal Process.}, vol.~68, pp. 6535--6546, 2020.

\bibitem{yan2022two}
Y.~Yan \emph{et~al.}, ``Two sides of the same coin: Heterophily and oversmoothing in graph convolutional neural networks,'' in \emph{2022 IEEE International Conference on Data Mining (ICDM)}.\hskip 1em plus 0.5em minus 0.4em\relax IEEE, 2022, pp. 1287--1292.

\bibitem{he2009learning}
H.~He and E.~A. Garcia, ``Learning from imbalanced data,'' \emph{IEEE Trans. Knowl. Data Eng.}, vol.~21, no.~9, pp. 1263--1284, 2009.

\bibitem{bradley1997use}
A.~P. Bradley, ``The use of the area under the {ROC} curve in the evaluation of machine learning algorithms,'' \emph{Pattern Recognit.}, vol.~30, no.~7, pp. 1145--1159, 1997.

\bibitem{kolaczyk2014statistical}
E.~D. Kolaczyk and G.~Cs{\'a}rdi, \emph{Statistical Analysis of Network Data with R}.\hskip 1em plus 0.5em minus 0.4em\relax Springer, 2014, vol.~65.

\bibitem{chung2014empirical}
J.~Chung, C.~Gulcehre, K.~Cho, and Y.~Bengio, ``Empirical evaluation of gated recurrent neural networks on sequence modeling,'' \emph{arXiv preprint arXiv:1412.3555}, 2014.

\end{thebibliography}

{\appendices
\section{Sensitivity analysis of threshold selection for graph sparsity and structural integrity}
\label{appendix}

In Table~\ref{tab:results1}, we evaluated the complexity metrics of edge density, $\eta(\mathcal{G})$, and edge entropy, $H(\mathcal{G})$, for the CPG across various thresholds and three data splits ($\mathcal{D}_{\text{train}}$), as defined in Section II.B of the main paper. Our threshold analysis covered values of $0.6$, $0.725$, $0.85$, and $0.975$, revealing that the $0.975$ threshold yielded the sparsest graphs, with edge densities between $0.02$ and $0.027$ and stable edge entropy values from $2.957$ to $3.389$. As thresholds decreased to $0.85$, edge density increased to $0.057$–$0.06$, and entropy rose slightly to $3.419$–$3.593$. This trend continued at the $0.725$ threshold, where edge density ranged from $0.078$ to $0.084$ with entropy values of $3.533$–$3.711$, resulting in a denser graph with a more complex edge distribution. At the lowest threshold, $0.6$, edge density reached $0.098$–$0.105$ with entropy ranging from $3.618$ to $3.806$. Across all methods—correlation, smoothness, and HGD-DTW—the sparsity and entropy metrics confirmed that decreasing thresholds generally increased both density and entropy. To allow fair comparisons, each graph’s edge entropy was standardized, enabling consistent assessment of performance differences between STP and CPG graphs.

Importantly, edge density never exceeded 10\% across all thresholds, even at $0.6$, ensuring that the graphs remained sparse enough to avoid excessive connectivity that could hide meaningful relationships.

\begin{table}[ht]
    \caption{Complexity metrics for the CPG using different thresholds. The metrics include: i) edge density, $\eta(\mathcal{G})$, and ii) edge entropy, $H(\mathcal{G})$, across three data splits for methods defined in SectionII.B of the main paper.}
    \centering
    \resizebox{\columnwidth}{!}{%
    \begin{tabular}{c|c|cc|cc|cc}
        \toprule
         & & \multicolumn{2}{c|}{Correlations} & \multicolumn{2}{c|}{Smoothness} & \multicolumn{2}{c}{HGD-DTW} \\
         Threshold & Split & $\eta(\mathcal{G})$ & $H(\mathcal{G})$ & $\eta(\mathcal{G})$ & $H(\mathcal{G})$ & $\eta(\mathcal{G})$ & $H(\mathcal{G})$ \\
         
        \midrule
        \multirow{3}{*}{0.975}
        & s1 & 0.026 & 3.389 & 0.02 & 3.356 & 0.027 & 2.983 \\ 
        & s2 & 0.021 & 3.366 & 0.02 & 3.356 & 0.021 & 2.957 \\ 
        & s3 & 0.020 & 3.356 & 0.02 & 3.356 & 0.020 & 2.960 \\ 
        
        \midrule
        \multirow{3}{*}{0.85}
        & s1 & 0.060 & 3.571 & 0.057 & 3.572 & 0.060 & 3.419 \\ 
        & s2 & 0.059 & 3.593 & 0.057 & 3.572 & 0.059 & 3.442 \\ 
        & s3 & 0.057 & 3.572 & 0.057 & 3.572 & 0.059 & 3.421 \\ 
        
        \midrule
        \multirow{3}{*}{0.725}
        & s1 & 0.084 & 3.711 & 0.078 & 3.673 & 0.083 & 3.574 \\ 
        & s2 & 0.079 & 3.680 & 0.078 & 3.673 & 0.080 & 3.555 \\ 
        & s3 & 0.078 & 3.673 & 0.078 & 3.673 & 0.081 & 3.533 \\ 
        
        \midrule
        \multirow{3}{*}{0.6}
        & s1 & 0.104 & 3.806 & 0.098 & 3.763 & 0.105 & 3.647 \\ 
        & s2 & 0.098 & 3.779 & 0.098 & 3.763 & 0.101 & 3.631 \\ 
        & s3 & 0.098 & 3.763 & 0.098 & 3.763 & 0.104 & 3.618 \\ 
        \bottomrule
    \end{tabular}%
    }
    \label{tab:results1}
\end{table}

In addition, we performed a visual analysis of the adjacency matrices for each method and threshold, shown in Figs.~\ref{f:AdjacencyMatrix_CPG}. 
The rows correspond to different threshold levels, ranging from $0.6$ to $0.975$ in increments of $0.125$, from top to bottom. 
These figures illustrate the impact of different thresholds on graph sparsity and structure across estimation methods. Integrating visual inspection with statistical analysis of edge density and entropy provides a deeper understanding of graph configurations, revealing specific patterns and irregularities that may not be captured by metrics alone. This approach enhances both the reliability and interpretability of our findings.

This same analysis of edge density and entropy was conducted for the graph estimation by time step (STG), and the results can be viewed in the following GitHub repository link: \url{https://github.com/oscarescuderoarnanz/XST-GCNN/blob/main/XST-GNN_Architecture/step2_graphRepresentation/graphComplexityMetrics.ipynb}. 
The conclusions regarding the threshold's impact were consistent, reinforcing that it has a limited effect on performance outcomes.

Ultimately, we selected the 0.975 threshold for all experiments in Section III.C.2) of the main paper as it best balances sparsity with essential structural detail for accurate modeling, showing a slight ROC-AUC performance advantage and validating its robustness for our study’s goals (see experiments at \url{https://github.com/oscarescuderoarnanz/XST-GCNN/tree/main/XST-GNN_Architecture/step3_GCNNs/Exp4_OthersExp_Performance_by_threshold}).

\begin{figure}[htpb]
\centering
    \begin{subfigure}
    \centering
    \includegraphics[width=0.15\textwidth]{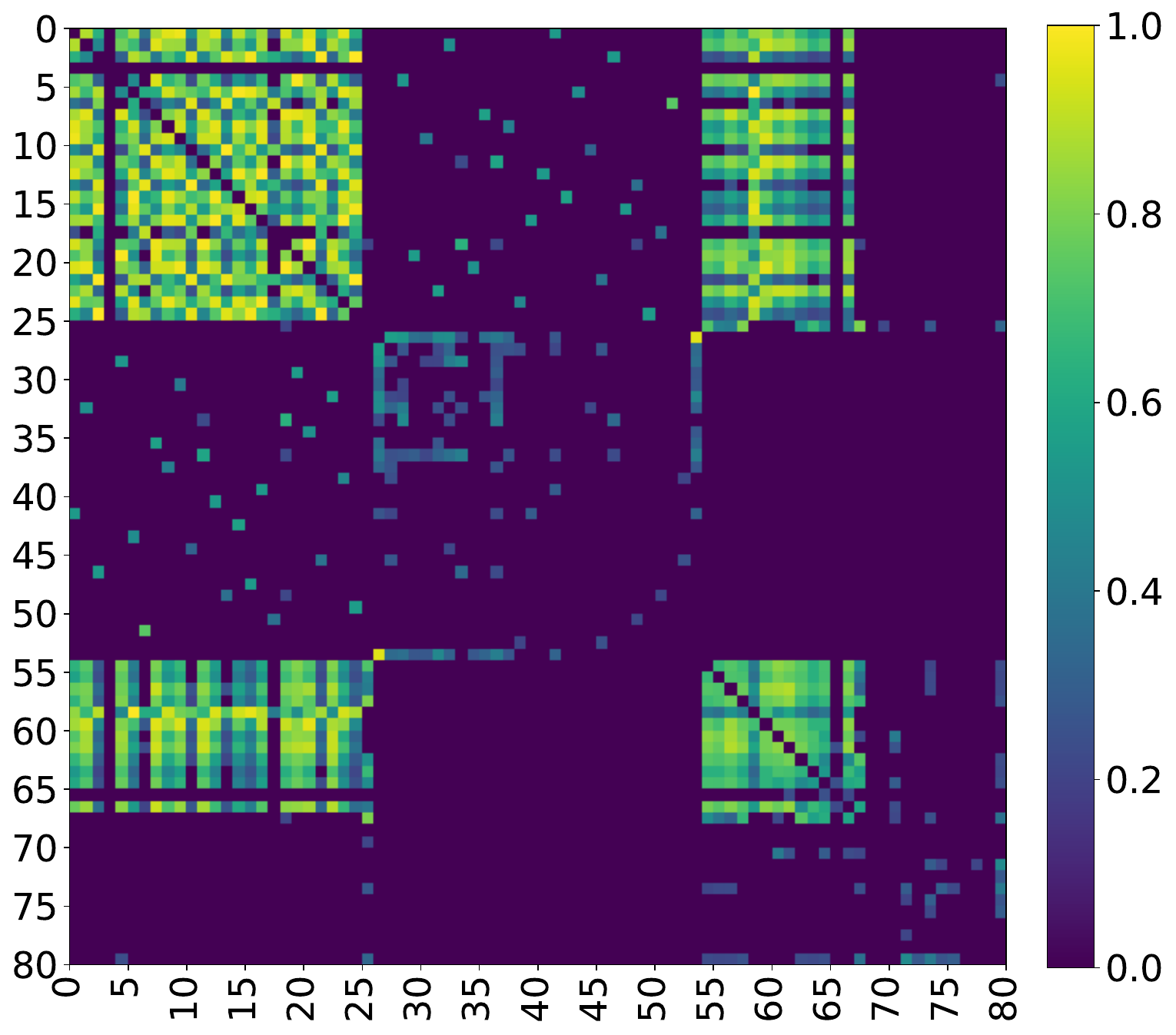}
    \end{subfigure}
    \begin{subfigure}
    \centering
    \includegraphics[width=0.15\textwidth]{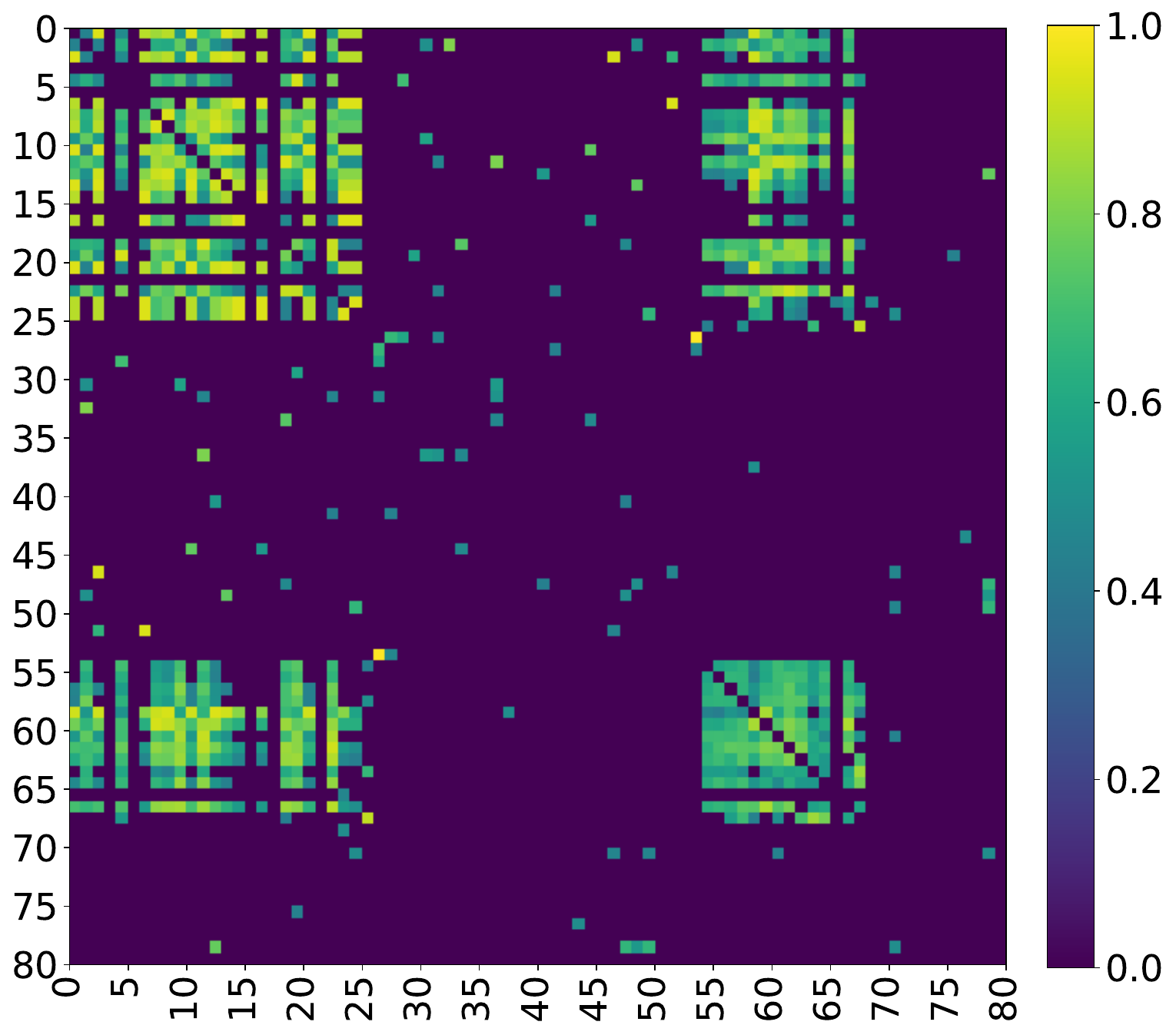}
    \end{subfigure}
    \begin{subfigure}
    \centering
    \includegraphics[width=0.15\textwidth]{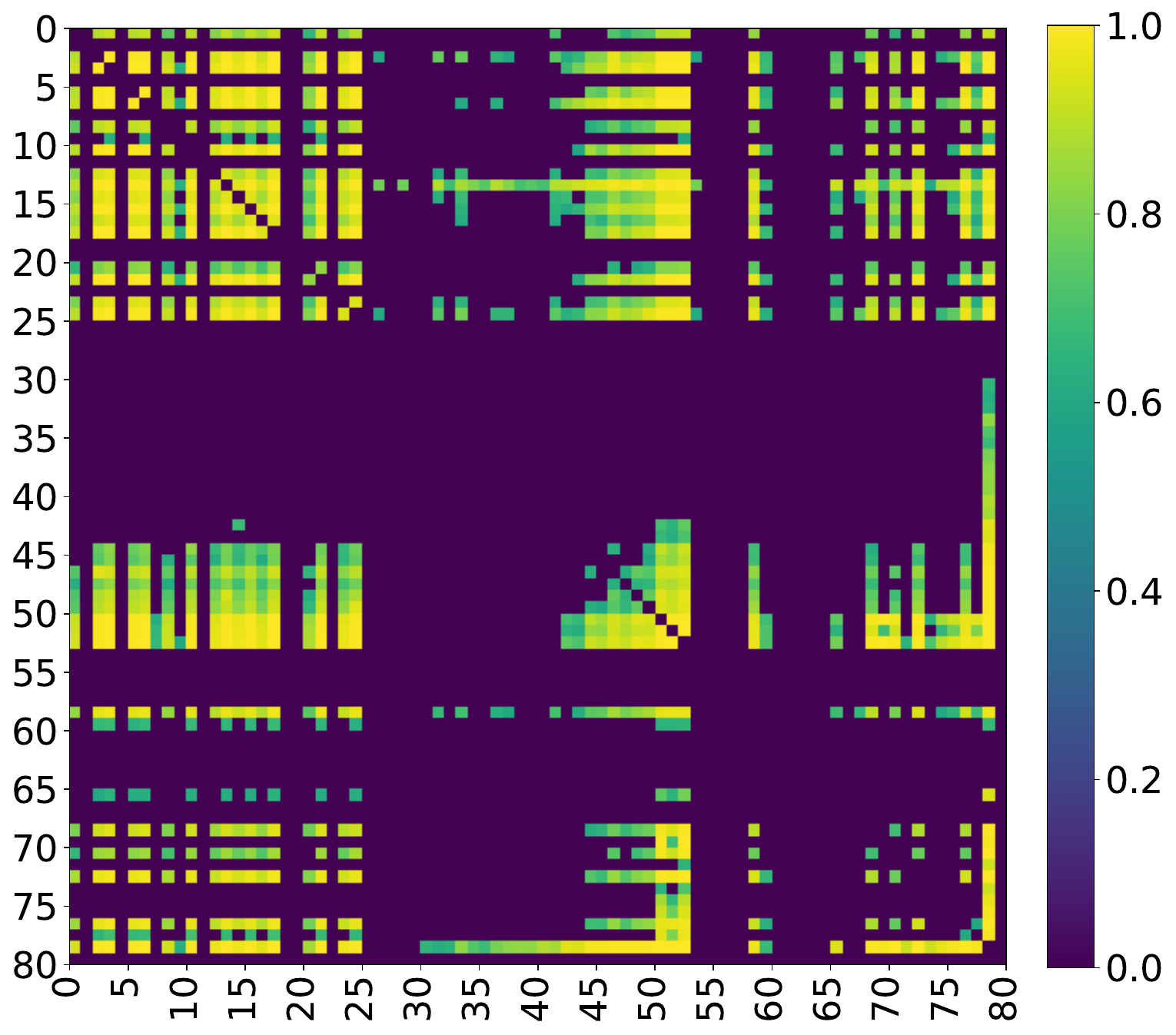}
    \end{subfigure}    
    
    \begin{subfigure}
    \centering
    \includegraphics[width=0.15\textwidth]{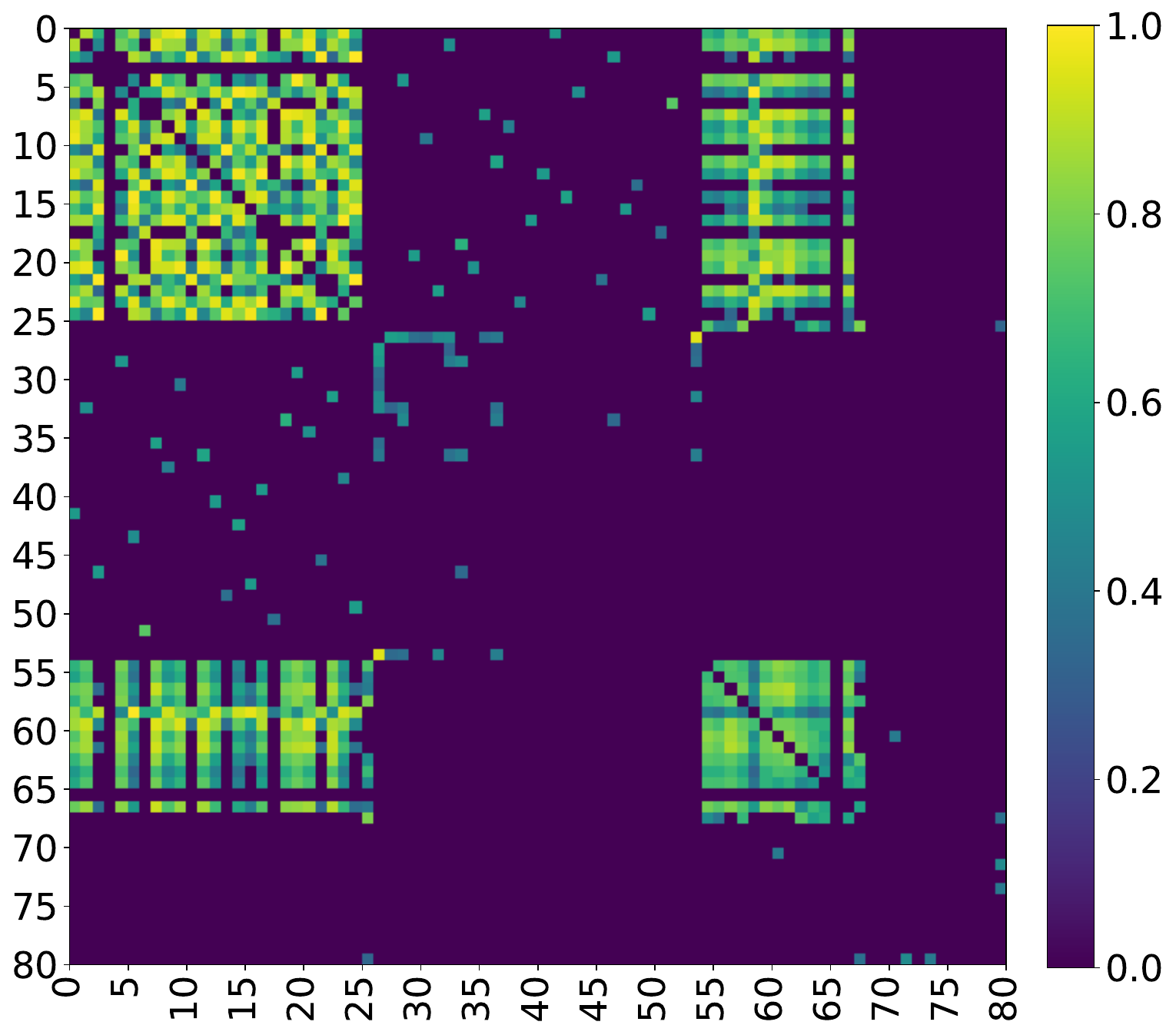}
    \end{subfigure}
    \begin{subfigure}
    \centering
    \includegraphics[width=0.15\textwidth]{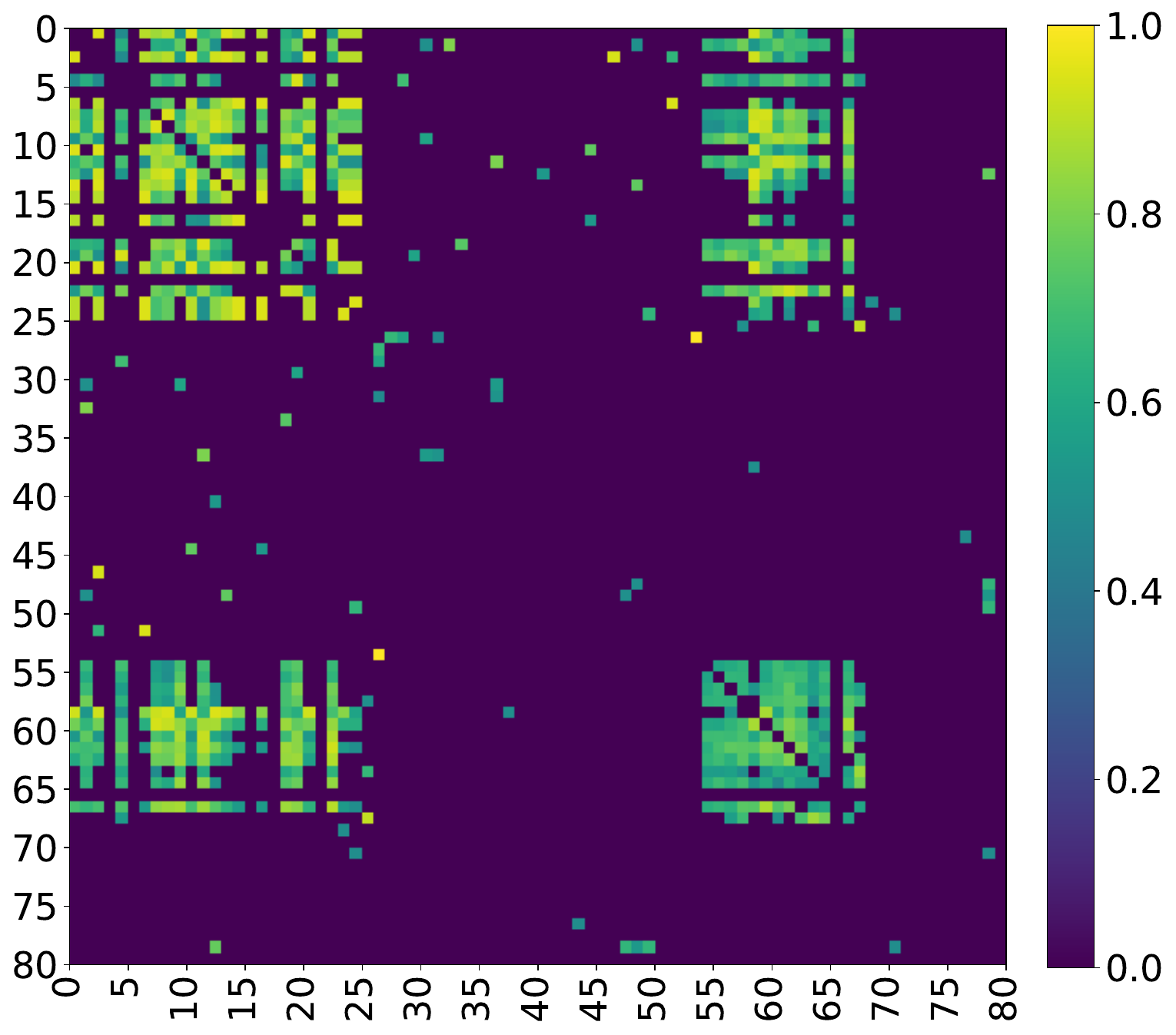}
    \end{subfigure}
    \begin{subfigure}
    \centering
    \includegraphics[width=0.15\textwidth]{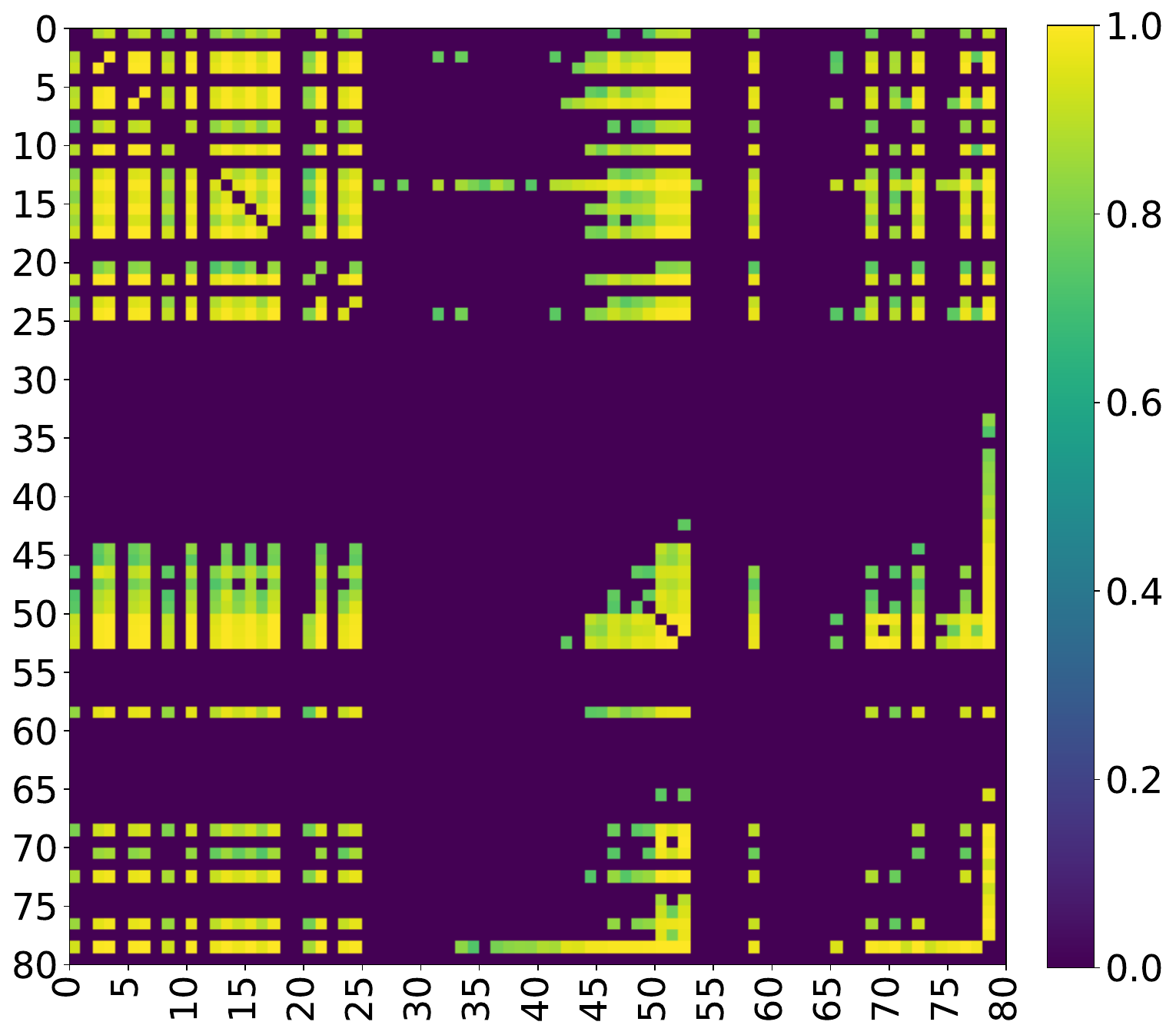}
    \end{subfigure}    
    
    \begin{subfigure}
    \centering
    \includegraphics[width=0.15\textwidth]{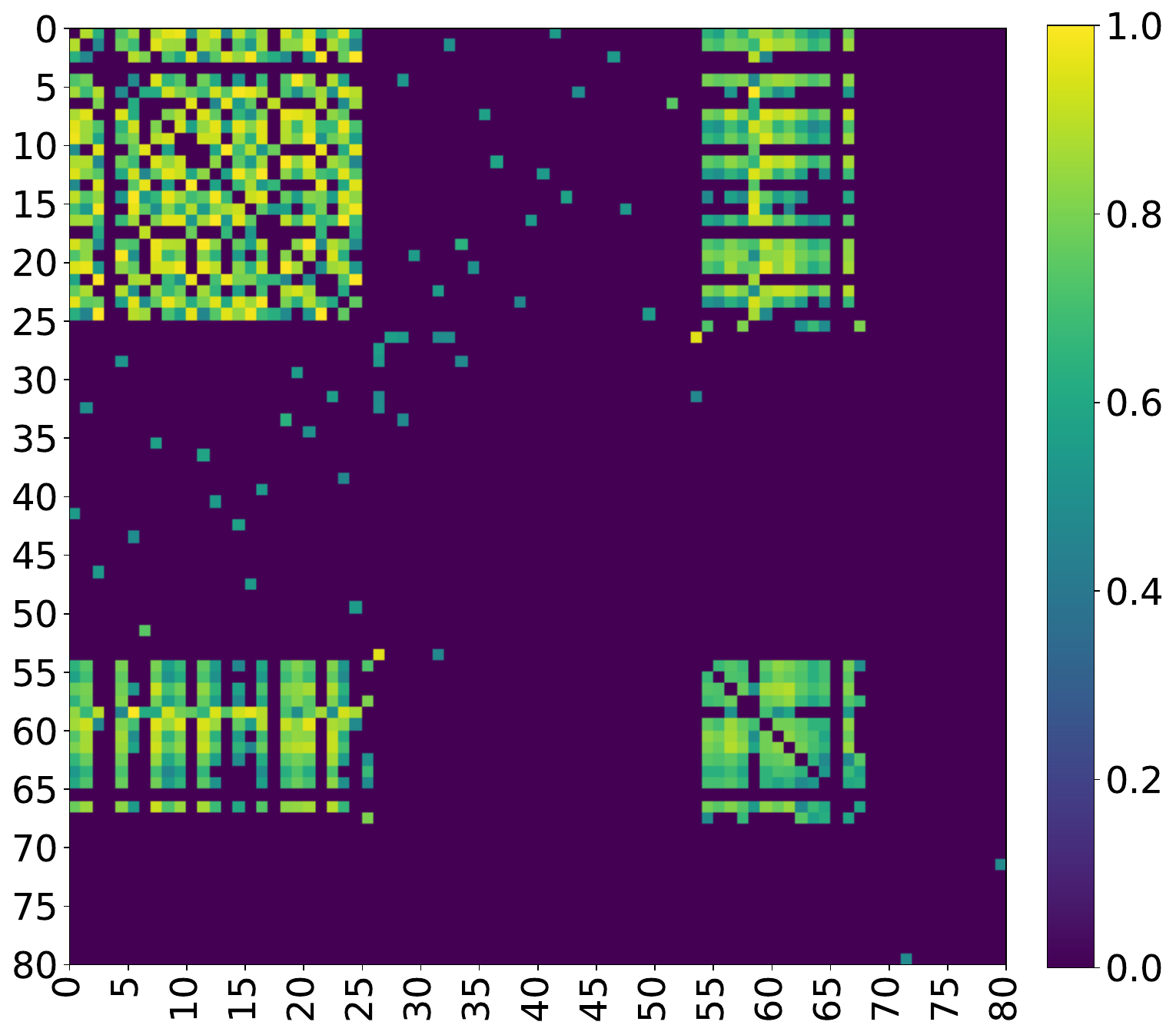}
    \end{subfigure}
    \begin{subfigure}
    \centering
    \includegraphics[width=0.15\textwidth]{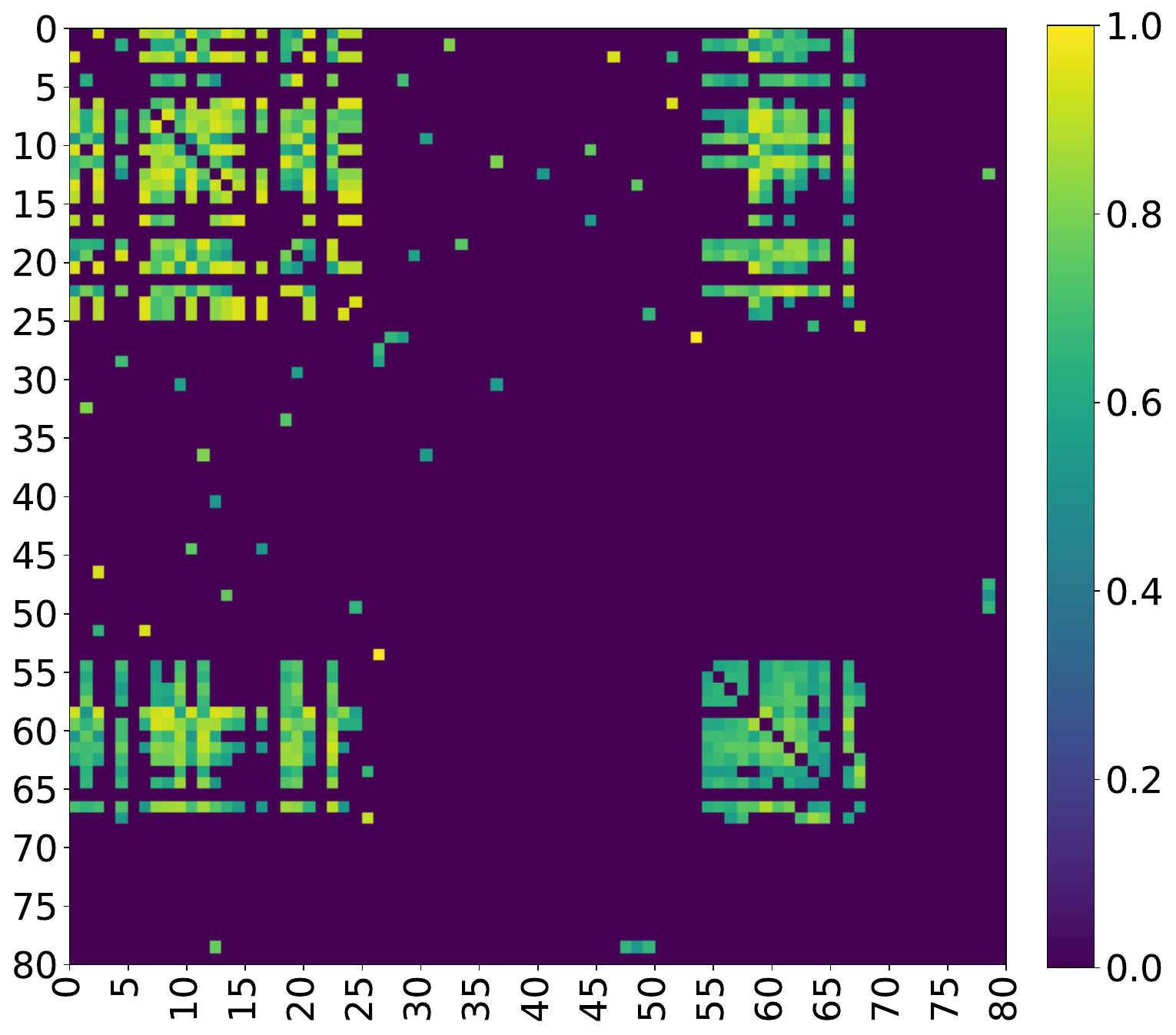}
    \end{subfigure}
    \begin{subfigure}
    \centering
    \includegraphics[width=0.15\textwidth]{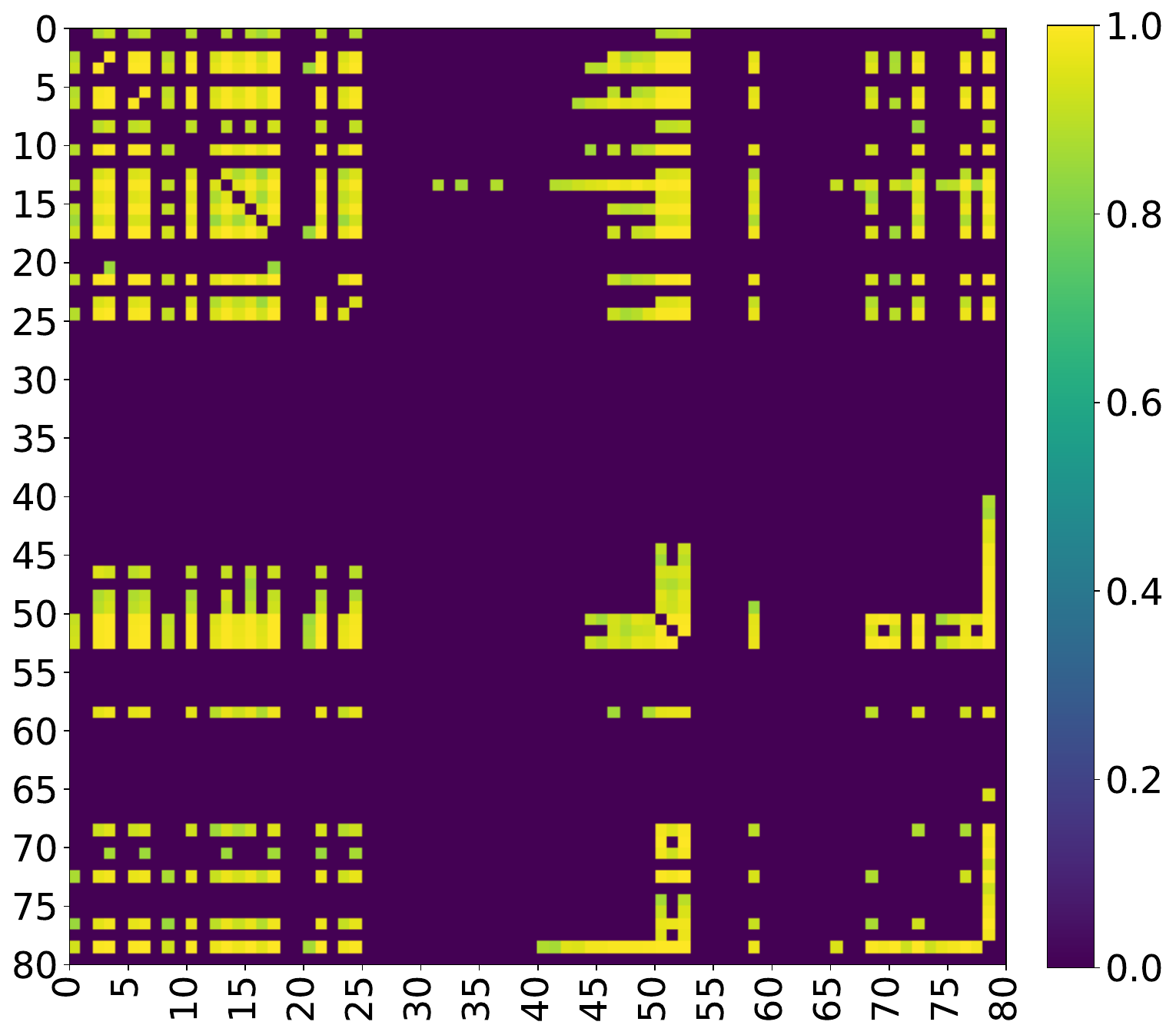}
    \end{subfigure}    
    
    \begin{subfigure}
    \centering
    \includegraphics[width=0.15\textwidth]{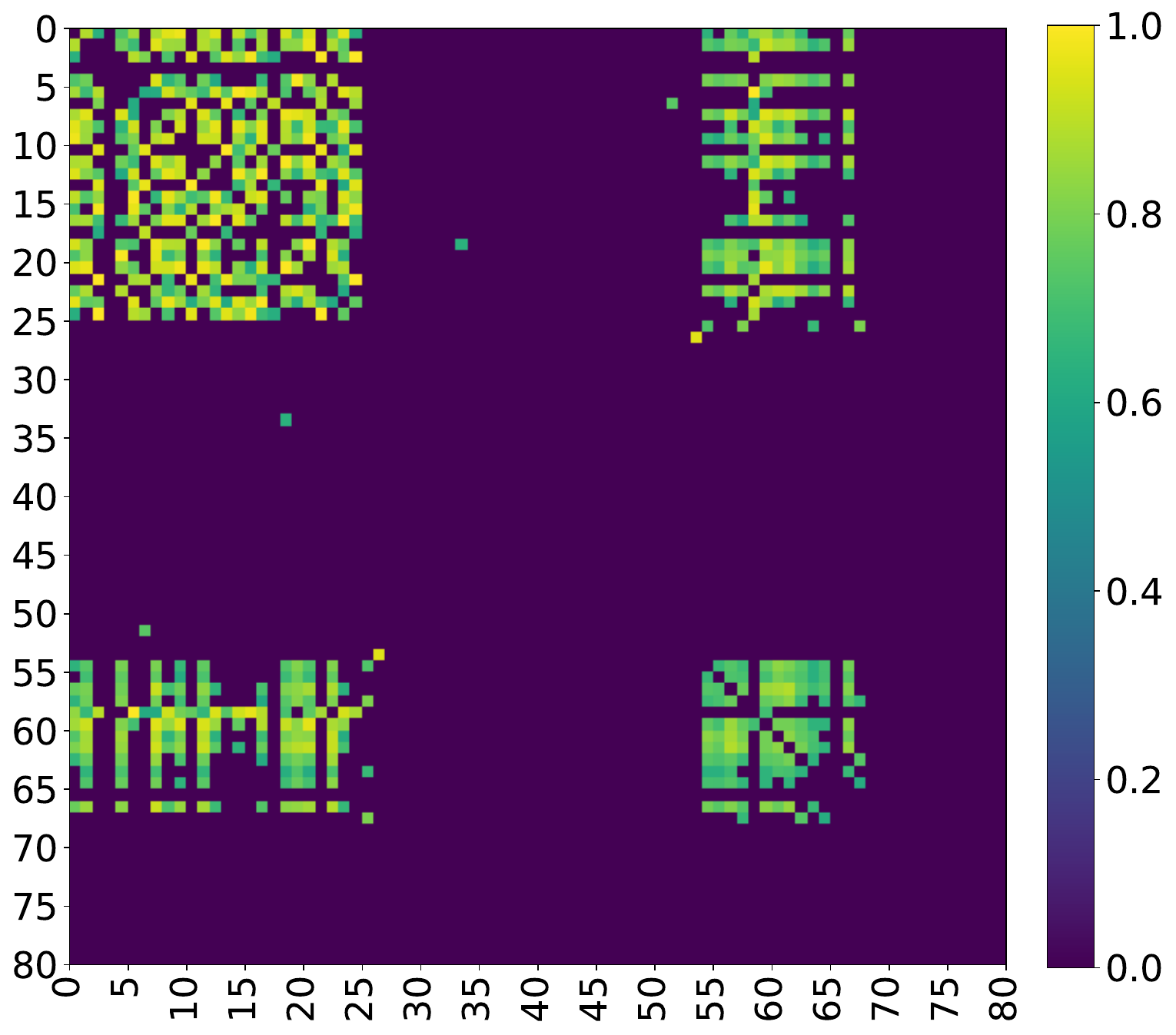}
    \end{subfigure}
    \begin{subfigure}
    \centering
    \includegraphics[width=0.15\textwidth]{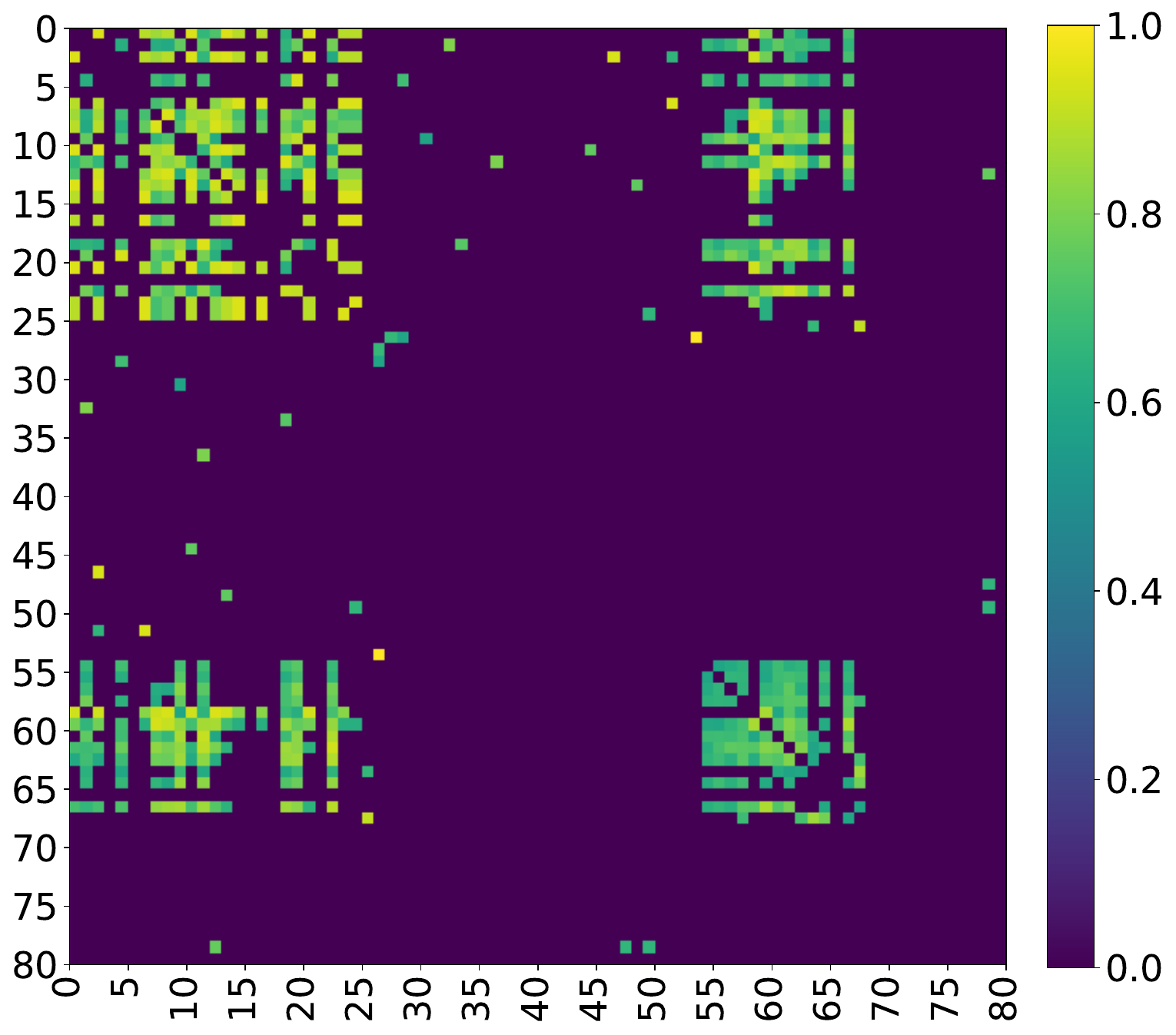}
    \end{subfigure}
    \begin{subfigure}
    \centering
    \includegraphics[width=0.15\textwidth]{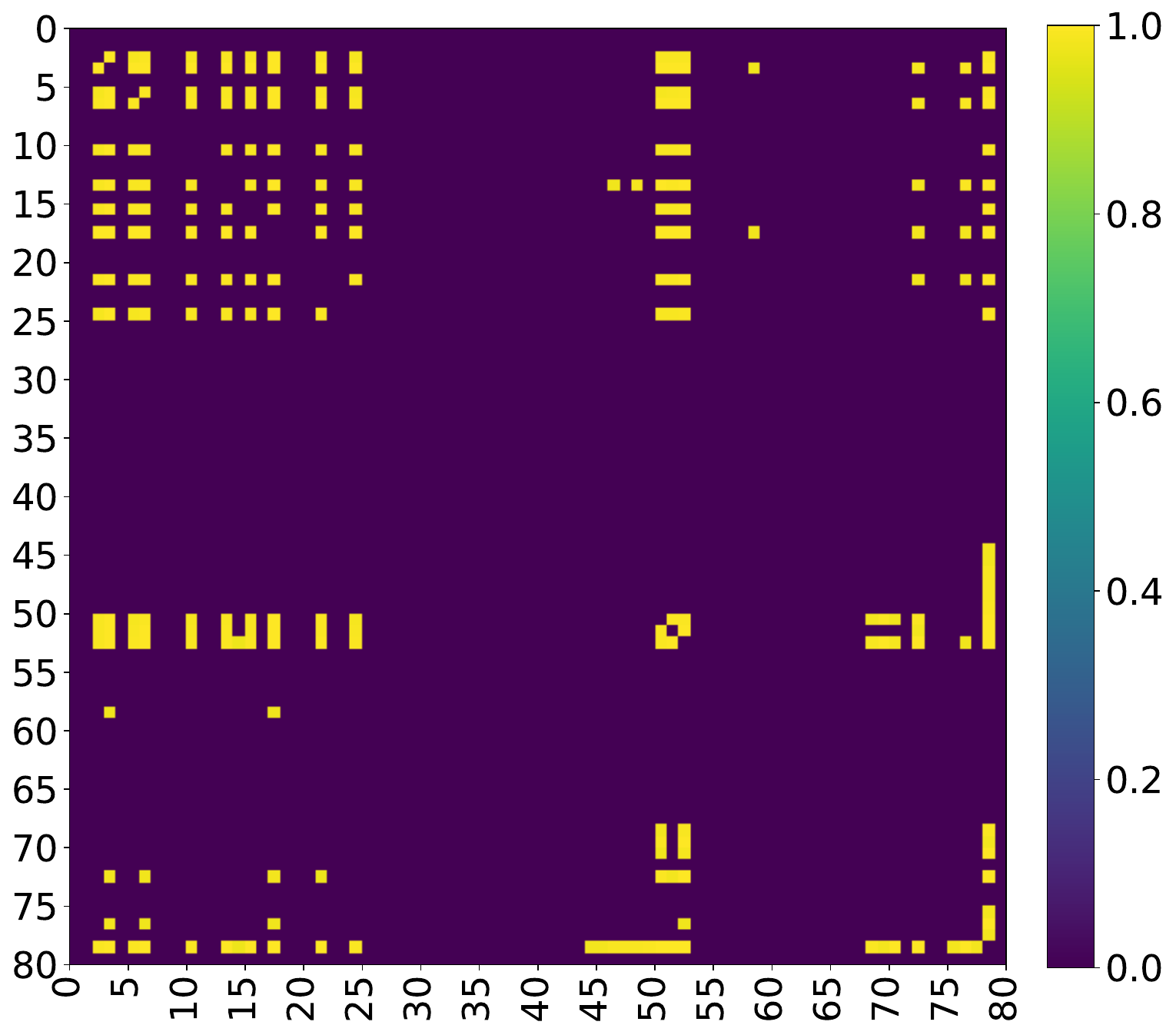}
    \end{subfigure}    
     
    \caption{Adjacency matrices representing the estimated graphs for the CPG using three different methods: correlation, smoothness, and DTW-HGW (from left to right in each row). The rows correspond to different threshold levels, ranging from 0.6 to 0.975 in increments of 0.125, from top to bottom.}
    \label{f:AdjacencyMatrix_CPG}
\end{figure}

\section{Pre-hoc Explainability for Node Importance in Graph Representations and Estimations}
\label{appendix2}

We conducted a comprehensive pre-hoc analysis of the estimated graphs, focusing on their temporal representations. This pre-hoc analysis, performed prior to model training, was essential for understanding the structure and interactions in the data after estimating the graphs and deciding on the most suitable graph representation. In a clinical context, interpretability is paramount: clinicians must not only trust model outcomes but also understand the interactions between variables, which can reveal critical insights into patient health and treatment efficacy. Graph-based representations offer a powerful and intuitive means to visualize these interactions, allowing clinicians to observe and analyze relationships among multiple variables over time, which can be crucial in supporting timely and informed decision-making. All generated graphs are available in the project repository for reference and reproducibility.\footnote{The complete set of estimated graphs can be found at \url{https://github.com/oscarescuderoarnanz/XST-GCNN/tree/main/XST-GNN_Architecture/step2_graphRepresentation}.}

In collaboration with clinical experts, we selected specific visualizations to illustrate the evolution of node importance over time, as depicted in Fig.~\ref{fig:stg}. Each visualization represents a time-specific graph estimated using HGD, the method associated with our best-performing architecture in terms of ROC-AUC (see Section III.C.2) of the main paper). This representation allows clinicians to track the importance and interactions of the 80 variables (nodes) over time, facilitating the identification of patterns and trends that inform clinical decision-making and provide insight into complex data relationships, helping unravel the underlying dynamics that impact patient outcomes. Node labels are provided in Fig.~\ref{fig:features}.

\begin{figure}[ht]
    \centering
	\includegraphics[width=1\columnwidth]{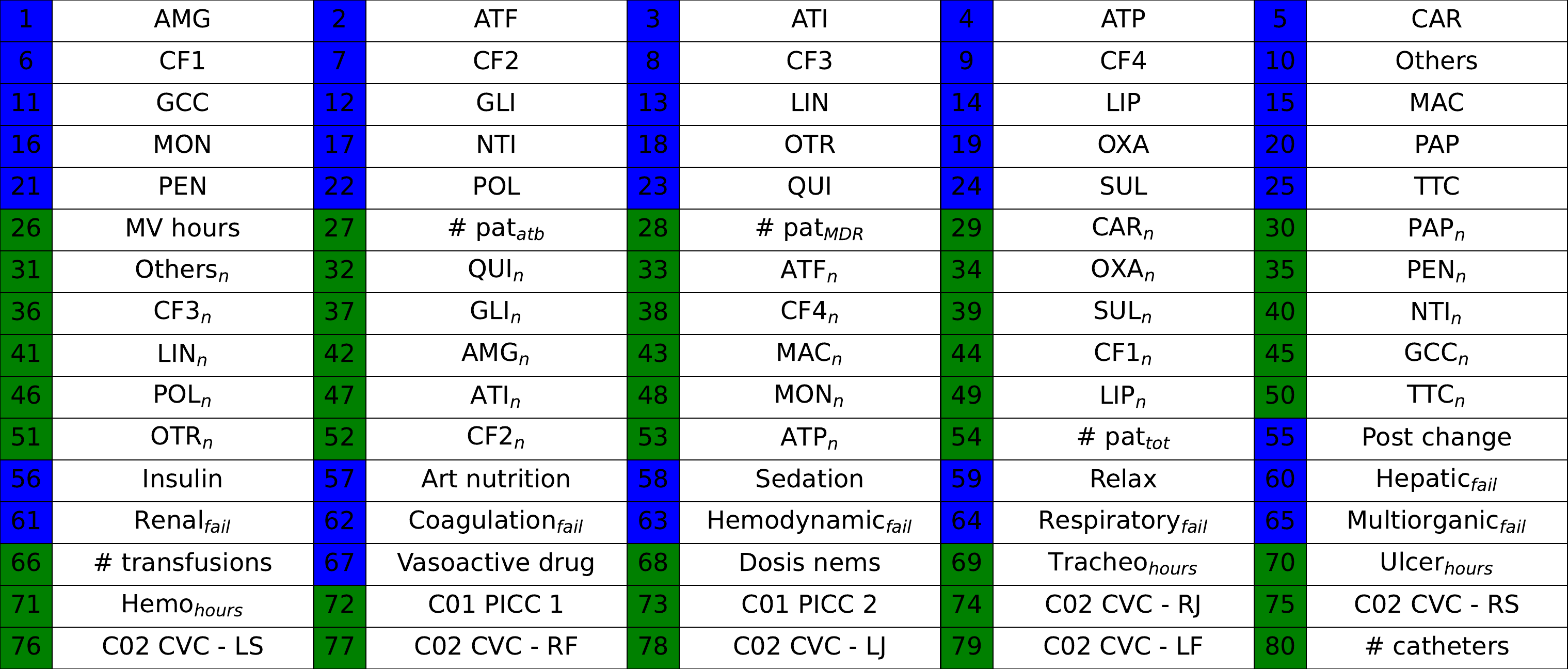}
    \caption{Overview of the variables utilized in this study for the real-world application case. The dataset comprises heterogeneous variables, categorized into binary variables (highlighted in blue) and continuous variables (highlighted in green). Each variable is annotated with its respective identifier for precise reference.}
	\label{fig:features}
\end{figure}

The temporal dynamics analysis in these graph representations reveals variable connectivity patterns that provide essential insights for patient monitoring and intervention strategies. At the initial time step (\( t_0 \)), core discrete variables form a central subgraph with a few continuous variables, while other discrete variables remain isolated. Continuous variables are generally well-connected, each having at least one edge to another variable. As time progresses to \( t_1 \), only the central subgraph remains connected, primarily composed of discrete variables (ATF, ATI, CF1, Others, LIN, NTI, PEN, SUL, and sedation) and a subset of continuous variables (OTR\textsubscript{n}, CF2\textsubscript{n}, and CO2 CVC - LJ). This pattern persists through \( t_2 \), with the addition of another continuous variable, TTC\textsubscript{n}, to the subgraph.

With further time progression, new interactions emerge. By \( t_7 \), we observe a re-establishment of connections among multiple variables, resembling the initial structure at \( t_0 \); several discrete variables become disconnected while all continuous variables maintain at least one connection. At later time points, previously isolated discrete variables remain unconnected, while linked variables in the central subgraph increase in both the number and weight of their connections.

This temporal visualization, integrated as a pre-hoc analysis within the XST-GCNN framework, provides clinicians with valuable insights into critical antibiotic families, co-patient exposure, and health status indicators that require further monitoring. By representing relationships between variables, this visualization enables clinicians to visually identify patterns, specifically those relationships that inform MDR prediction. Recognizing these patterns allows clinicians to make informed decisions, such as implementing pre-emptive isolations or reducing contact between patients receiving similar treatments. These actions help mitigate transmission risks, improve patient outcomes, and contribute to a safer and more efficient ICU environment.

\begin{figure*}[ht]
    \centering
	\includegraphics[width=0.725\columnwidth]{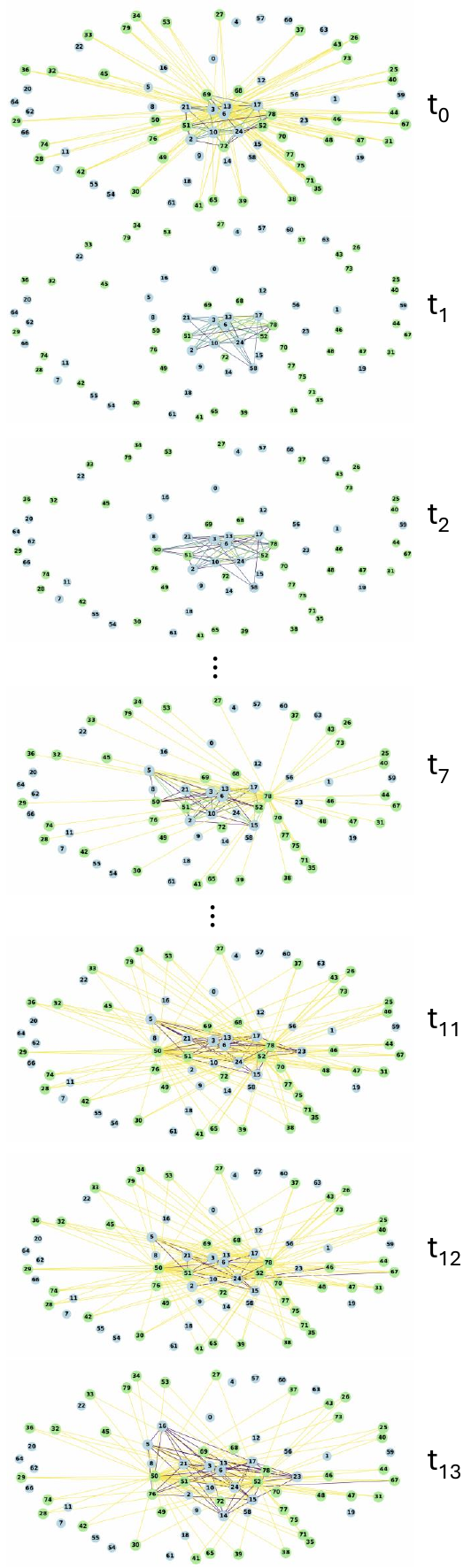}
    \caption{Temporal evolution of estimated graphs using the HGD method. Each row represents a graph at a specific time step, illustrating the changing connectivity patterns between variables over time. Nodes correspond to distinct clinical variables, with edges indicating the relationships and dependencies captured at each graph temporally. This visualization enables analysis of the dynamics in variable importance and interaction, providing valuable insights for MDR prediction.}
	\label{fig:stg}
\end{figure*}

\end{document}